# Scalpel and Brush: Pruning and Targeted Weight Editing of a Small Language Model by a Single LLM Agent

**Sergei Koltcov[1], Vera Ignatenko[1], Anton Surkov[1], Rabie Abdultawab Ramadan[2]**

[1] Laboratory for Social and Cognitive Informatics, Higher School of Economics, St. Petersburg, Russia

[2] Department of Information Systems, University of Nizwa, Nizwa, Oman

## Abstract

It has recently been shown that an LLM can itself edit the weights of a language model, yet it remains unclear which part of such a result is due to the agent and which part is reproduced by a random edit of the same shape, by an instruction in the prompt, or by a standard procedure without an LLM. We unify repair and pruning as sparse weight edits to optimize a metric we specify, using BLEU as an example: three tasks — targeted weight repair ("brush"), structured MLP neuron pruning ("scalpel", zeroing pruned weights), and layer removal with repair — are assigned to a DeepSeek Harness agent powered by DeepSeek-V4.1-Flash, with up to 50 dev evaluations. We study Qwen2.5-Math-1.5B-Instruct in three independent runs per task, with final evaluation on two hidden splits. On ordinary differential equations, changing at most 0.0084% of weights raised test BLEU by 24–32% on second-order equations and by a factor of 1.8–2.5 on equations with polynomial right-hand sides. An independent substitution check showed that higher BLEU reflects notation alignment rather than improved correctness. Masks that remove 12.5–15% of MLP neurons shrink the model by 9.4–11.2% with mean BLEU loss of at most 0.01 on both hidden splits, whereas standard pruning with equal shares per layer failed the constraint at all tested shares. On seven-class GoEmotions classification, edits raise macro-F1 from 0.100 to 0.396–0.451 on test and from 0.175 to 0.455–0.476 on holdout. A ridge-based classifier written into agent-selected weight rows yields macro-F1 of 0.424 and 0.464. Pruning search with the same evaluation limit removes 15.9% of neurons versus 14–21.4% for agent runs passing the hidden check. Standard procedures without an LLM reproduce a substantial part of the performance once given the location and form of intervention found by the agent; a comparison of agentic and non-agentic search remains open.

## 1. Introduction

The behavior of a trained language model is changed by fine-tuning, including low-rank fine-tuning (LoRA [31]), by activation steering without changing the weights — steering vectors [5–7] — and by direct editing of identified weights, as in fact editing with ROME and MEMIT [2, 3]; for mathematical models, steering was studied in our MathSteering project [36], from which the data and the metric of the present work are taken. Pruning, at first sight, solves the opposite task — remove as many weights as possible while preserving quality — but it is structured in the same way: weights or neurons are ranked by their contribution to the model error [8–16], except that pruning removes the least important ones, while editing changes the most important ones. It is therefore natural to consider both operations over a single importance distribution; these directions are discussed in more detail in Section 2.

It has recently been demonstrated that direct weight editing can be performed by an LLM itself. In "Sparks of Weight-Space Programming" [1], one of the leading language models studied the weights of Qwen3.5-0.8B, found the cause of looping during reasoning and changed 64 columns of `down_proj` in each of layers 15 and 19 — 131,072 weights in total, about 0.016% of the model; accuracy on MATH-500 rose from 7.9 to 48.1%. The problems on which the agent searched for the edit are separated from the final tests, and an instruction-following evaluation (IFEval) is reported in addition to mathematics. There is no comparison with random edits of the same size in that experiment, so it remains open which part of the effect is due to the found direction and which part to the mere fact of intervention.

Hence the question addressed in this work: **what does an LLM agent that is given both compression and repair of a model in a single setting find by itself, and which part of the result is reproduced by standard procedures**. We study this question on Qwen2.5-Math-1.5B-Instruct [32] and two tasks: solving ODEs, where the target metric BLEU measures the notation of the answer, and emotion classification of short texts, where macro-F1 directly measures classification quality. The agent — a digital twin of the researcher in the DeepSeek Harness environment with the deepseek-v4.1-flash engine — has full access to the weights and training data but evaluates its edits only through evaluation scripts with a limited number of calls. The results of the correctness check and the evaluations on held-out splits are not available to the agent: we use them to judge independently what an edit has actually changed. Every result of the agent is compared with random edits of the same shape, with a prompt instruction and, in the second domain, with procedures without an LLM. The work is a proof of concept: we study the feasibility of an autonomous workflow and its observable consequences rather than propose a universal optimizer for pruning or weight editing.

The main results are as follows:

1. Three tasks — repair ("brush"), neuron pruning ("scalpel") and layer removal with repair — are posed uniformly as a sparse edit $W \to W + \Delta$ with a common agent, data, evaluation procedure and limit of dev calls; the tasks differ only in the objective and in the form of $\Delta$.
2. A protocol for verifying the agent's edits is proposed: random controls of the same size, a correctness check whose results are not reported to the agent, repeated runs, comparison of the agent's explanations with measurements, and a single evaluation of the final variants on two held-out splits.
3. A common importance scale of MLP neurons is built, together with a map of which neurons the agent modifies during repair and which it removes during pruning (Fig. 2).
4. It is shown that the agent itself discovers the dependence of BLEU on notation and raises the metric by bringing the notation of formulas to the style of the references; the share of correct solutions does not change, and a prompt instruction does not produce the same effect.
5. In the second domain it is shown that the same procedure yields substantive repair: in all three runs a weight edit raises macro-F1 several-fold on both held-out splits, and the gain is not explained merely by a higher share of valid answers. The scalpel on this task gives very different results from run to run.
6. The contribution of the agent is compared with that of standard procedures. The agent itself chooses the location and form of the edit, the layer-wise distribution of pruning and the controls; once these decisions are made, procedures without an LLM reach comparable quality.

The paper is organized as follows. Section 2 reviews related work, Section 3 describes the data, Section 4 the model, Section 5 the problem setting, the agent and the metrics, and Section 6 the baselines and controls. The results are presented in Section 7, the discussion and limitations in Sections 8 and 9. The Appendix contains a detailed literature review, the definitions of the controls and the course of the scalpel runs.

## 2. Related Work

**Weight editing and activation steering.** Knowledge-editing work assumes that model behavior is localized in a small part of the weights: ROME [2] uses causal tracing to locate factual knowledge in the MLPs of middle layers and replaces a fact with a rank-one edit, and MEMIT [3] extends the approach to thousands of facts; an edit is considered successful if it changes the target behavior and leaves the rest intact. Steering vectors [5–7] switch behavioral properties on and off by adding a direction to the residual stream without changing the weights. The two approaches are related: a column $d_i$ of the $down_proj$ matrix is the direction that a neuron adds to the stream in proportion to its activation, so editing such columns implements a conditionally activated steering vector; Yin et al. [34] proposed a close construction with a trigger. The search for an edit is automated with reinforcement learning [4] or, as in [1], delegated to another LLM that reads the weights and formulates hypotheses itself. We complement the latter setting with random edits of the same size and a correctness check that separates a change in the notation of the answer from a change in the solution.

**Pruning.** Pruning methods assign an importance score to weights or structures and remove the least important ones: from the second derivative of the loss [8] and weight magnitude [9] to Taylor expansion [10], SNIP [11] and learned masks [12–14]. For large language models, methods that need only a small calibration set have become dominant: SparseGPT [15] and Wanda [16], the structured LLM-Pruner [17] and SliceGPT [18], removal of whole blocks in ShortGPT [19] and SLEB [21], and removal of deep layers with recovery by QLoRA fine-tuning [20]. Two observations from these works matter for us: the quality of pruning depends not only on the criterion but also on the calibration and the layer-wise allocation of the share, which are usually chosen by a human; and pruning can not only degrade but also repair a model — removing individual attention heads improves reasoning accuracy, including for Qwen2.5-Math-1.5B-Instruct [22], and removing individual neurons improves generalization [23].

**A physical view of pruning.** Importance criteria answer the question of what to remove but not how much. In our work [24, 25], the free energy of activations predicts the accuracy of a network after pruning, and pruning itself is described as a phase transition with a critical share that can be estimated from the activations of the original network without repeated runs. This leads to the expectation that agent-driven pruning also has a critical share. The drop in correctness for the agent's largest mask (Sections 7.3, 7.5) is consistent with it, but it does not prove a general threshold: masks of different shares also differ in neuron composition and layer-wise allocation.

**LLM agents and the position of this work.** Agents already carry out research from hypothesis to report [29, 30] and are used for compression: they assign layer-wise pruning ratios [26], search for compression strategies within a given procedure [27], and choose the layers to remove and the recovery methods [28]. Our agent differs in having open access to weights and activations: it writes its own analysis code and builds sparse edits within an evaluation limit, and we test repair, neuron pruning and layer removal on the same model and with the same metric. Representing both operations as an edit $W \to W + \Delta$ is an experimental framework rather than a new optimization algorithm; we are interested in which edits an agent will find when it is free in its analysis but limited in the number of evaluations, what explains their effect, and whether the result transfers to data the agent has not seen. A detailed review is given in the Appendix (A.1).

## 3. Data

**Ordinary differential equations.** We use a synthetic ODE dataset from the MathSteering project [36]. Each example consists of an equation in LaTeX and a reference general solution with arbitrary constants. Equation types:

- **ode2** — linear inhomogeneous second-order equations with constant coefficients $a\,y'' + b\,y' + c\,y = f(x)$, where $f$ is an exponential, trigonometric or hyperbolic function or a combination of them. The solution contains constants $C_1, C_2$. Example: $-y'' + 4y' - 5y = \sinh 2x$, answer $y = (C_1 \sin x + C_2 \cos x)\, e^{2x} - \frac{9}{17}\sinh 2x - \frac{8}{17}\cosh 2x$.
- **poly** — equations $y' = P(x)$ with a polynomial right-hand side; the solution is a polynomial plus a constant. Example: $y' = 2305 + 6704x + 8694x^2 - 14640x^3 + 450x^4$, answer $y = 2305x + 3352x^2 + 2898x^3 - 3660x^4 + 90x^5 + C$.
- **separable** — first-order equations $y' = F(x)$ whose right-hand side consists of products and quotients of elementary and inverse trigonometric functions; the solution is an antiderivative plus a constant. Example answer: $y = \mathrm{arccot}x \cdot \sin x + \ln 2x \cdot \arccos x + C$.

The experiments use 645 problems: 150 form the dev split and 495 the pool. The dev split was sampled at random (50 problems of each type), the pool consists of the remaining cleaned examples; test and holdout are separate files stored outside the project directory.

| Split | ode2 | poly | separable | Total | Role | Visible to the agent |
|---|---|---|---|---|---|---|
| pool | 127 | 211 | 157 | 495 | analysis, calibration, fitting of edits | yes, fully |
| dev | 50 | 50 | 50 | 150 | evaluation of edits, ≤ 50 calls per run | only through the evaluation script |
| test | 119 | 60 | 69 | 248 | final evaluation | no |
| holdout | 150 | 150 | 150 | 450 | a single evaluation at the end | no |

**Reference style.** The same formula can be written in LaTeX in different ways, and BLEU compares precisely the notation, so the style of the references matters. The ode2 and poly references were printed to LaTeX by the symbolic mathematics library sympy and are written uniformly: subscripts and superscripts always in braces (`C_{1}`, `x^{2}`), function arguments in `\left( … \right)`. The model writes differently — for example, `C_1` instead of `C_{1}` (in 47 of 50 ode2 answers on dev) — and loses part of BLEU for this reason alone even when the answer is correct. The separable references are written in a different style (`\arcctg`, `\cdot`, `\ln(2 \cdot x)`); on the separable references in holdout see Section 7.5.

**Emotion classification.** We use the open GoEmotions dataset [35] — Reddit comments annotated with 27 emotions and a "neutral" label (median length 12 words). The emotions are grouped into Ekman's six groups — anger, disgust, fear, joy, sadness, surprise — and the neutral class is added, giving 7 classes in total. Only single-label texts are used; duplicate texts and texts shorter than 3 or longer than 40 words are removed. All splits are class-balanced (seed 0):

| Split | Per class | Total | Source (GoEmotions part) | Visible to the agent |
|---|---|---|---|---|
| pool | 300 | 2100 | train | yes, fully |
| dev | 50 | 350 | train | only through the evaluation script |
| test | 70 | 490 | test | no |
| holdout | 60 | 420 | dev | no |

The splits do not overlap in texts. The official test and dev parts of the dataset are stored outside the project directory; the agent is forbidden to download data, which is verified by an audit of its logs.

## 4. Model

We study **Qwen2.5-Math-1.5B-Instruct** [32], a mathematical version of Qwen2.5 fine-tuned to solve problems with step-by-step reasoning. The architecture is a decoder-only transformer:

| Parameter | Value |
|---|---|
| Layers (decoder blocks) | 28 |
| Hidden size $d$ | 1536 |
| Attention | 12 query heads, 2 key/value heads (GQA), head size 128, RoPE ($\theta = 10^4$) |
| MLP | SwiGLU, 1536 → 8960 → 1536 |
| Normalization | RMSNorm before attention and MLP, $\varepsilon = 10^{-6}$ |
| Vocabulary | 151,936 tokens; the embedding matrix is tied to the output layer |
| Number of parameters | 1.54 B, of which MLP 1.16 B (75%), embeddings 0.23 B |
| MLP neurons | $28 \times 8960 = 250\,880$ |

The **residual stream** is a vector of dimension $d = 1536$ that holds the current representation of each token and is passed from layer to layer. Layers do not replace this vector but add their corrections to it: the attention block adds information from other tokens, the MLP its own correction. The name comes from the residual connections $h \to h + f(h)$ that bypass each sub-block. The next-token probabilities are computed from the residual stream at the output of the last layer. Since all layers write into the same space, a direction in it can be interpreted as a feature, and adding a vector to the stream as an intervention in what all subsequent layers "see". Steering vectors and edits of *down_proj* columns are based on this.

Block $\ell$ transforms the residual stream $h_\ell$ as $h'_\ell = h_\ell + \mathrm{Attn}_\ell(\mathrm{RMSNorm}(h_\ell))$, $h_{\ell+1} = h'_\ell + \mathrm{MLP}_\ell(\mathrm{RMSNorm}(h'_\ell))$, where

$$\mathrm{MLP}_\ell(x) = W^\ell_{\mathrm{down}}\, a(x), \qquad a(x) = \mathrm{SiLU}\big(W^\ell_{\mathrm{gate}} x\big) \odot \big(W^\ell_{\mathrm{up}} x\big) \in \mathbb{R}^{8960}.$$

Neuron $i$ of layer $\ell$ adds the vector $a_i(x)\, d_i$ to the residual stream, where $d_i = W^\ell_{\mathrm{down}}[:, i]$. The **logit** of token $t$ is an unnormalized score that the next token is $t$; probabilities are obtained from logits by softmax. The logit equals $e_t^\top \mathrm{RMSNorm}(h_{28})$, where $e_t$ is a row of the embedding matrix: in this model the same matrix serves as the input embeddings and as the output layer.

**Generation.** Weights are stored in bf16. Decoding is greedy: at each step the most probable token is chosen, so the answer is deterministic, and any weight change that shifts even one choice changes all subsequent text. The answer length is limited to 1500 tokens; problems are processed in batches of 12. The prompt follows the MathSteering template in the model's chat template: "Solve the following differential equation step by step, then give the final answer in LaTeX format. Give your final answer in exactly the form: *y = <LaTeX expression>.*", followed by the equation. The answer is the content of the last *\boxed{}* (or of the last line *y = …*).

**Baseline quality** [dev]:

| Type | BLEU | Parsed by sympy | Correct | Truncated at 1500 tokens | Mean length, tokens |
|---|---|---|---|---|---|
| ode2 | 0.461 | 96% | 88% | 4% | 1104 |
| poly | 0.363 | 100% | 92% | 0% | 606 |
| separable | 0.116 | 18% | 0% | 8% | 1043 |

The model almost always solves ode2 and poly correctly, but its BLEU is moderate; it almost never solves separable in closed form and usually leaves the answer with an integral.

## 5. Problem Setting

### 5.1. Weight editing

All tasks are formulated as a parameter edit $W \to W' = W + \Delta$. The number of changed weights is denoted $\| \Delta \|_0 = \#\{(i,j)\colon \Delta_{ij} \neq 0\}$. An edit is applied as $W' = \mathrm{bf16}(\mathrm{fp32}(W) + \mathrm{fp32}(\Delta))$; with a zero edit the model's answers coincide with the original ones character for character.

**Pruning neuron** $i$ of layer $\ell$ means setting $\Delta = -W$ on row $i$ of $W^{\ell}_{\mathrm{gate}}, W^{\ell}_{\mathrm{up}}$ and on column $i$ of $W^{\ell}_{\mathrm{down}}$ ($3 \cdot 1536$ weights).

A **scalpel mask** is the result of the pruning task: a binary matrix $m \in \{0,1\}^{28\times 8960}$, where $m_{\ell,i} = 1$ means that neuron $i$ of layer $\ell$ is pruned (stored as JSON *{"<Layer>": [neuron indices], …}*). The agent passes the mask to the evaluation script, which zeroes the weights of the marked neurons and measures BLEU. With a mask, the MLP output is

$$\mathrm{MLP}_\ell(x) = \sum_{i=1}^{8960} (1 - m_{\ell,i})\, a_i(x)\, d_i,$$

the remaining weights are unchanged, and the pruned neurons can be physically removed from the matrices. A mask is described by the **share** of pruned neurons $\rho(m) = \frac{1}{250\,880} \sum_{\ell,i} m_{\ell,i}$ and by the **layer-wise allocation** — the shares $\rho_\ell = \frac{1}{8960} \sum_i m_{\ell,i}$.

**Removing layer** $\ell$ means replacing the block with the identity map $h_{\ell+1} = h_\ell$ (attention, MLP and both normalizations of the block are excluded).

### 5.2. Metrics

**The target metric on ODEs is BLEU** [33]: the degree of agreement between the notation of the model's final answer $\hat{a}$ and the reference $a$. Before comparison, the prefix *y(x) =* is removed from the reference, spaces and the commands *\left*, *\right* are removed from both strings, and *\cdot* and *\times* are replaced by *. The strings are then split into elementary parts — numbers, letters, LaTeX commands and symbols — and sequences of $n$ consecutive parts ($n$-grams) are compared:

$$\mathrm{BLEU}(\hat{a}, a) = \mathrm{BP} \cdot \exp\left(\sum_{n=1}^{4} 1/4 \log p_n\right), \qquad \mathrm{BP} = \min\left(1,\ e^{\,1-|a|/|\hat{a}|}\right),$$

where $p_n$ is the share of $n$-grams of the answer that occur in the reference and the factor BP penalizes a too-short answer; standard smoothing is used so that the metric does not vanish when there are no long matches. The implementation is the same as in MathSteering (function sentence_bleu of the nltk library). For an equation type $t$ the metric is the mean over problems, $\overline{\mathrm{BLEU}}_t = \frac{1}{N_t}\sum_{i\in t} \mathrm{BLEU}\,(\hat{a}_i, a_i)$, and its change is $\Delta\mathrm{BLEU}_t = \overline{\mathrm{BLEU}}_t(W + \Delta) - \overline{\mathrm{BLEU}}_t(W)$.

**Solution correctness** is a control characteristic. BLEU compares notation: a correct solution in a different form may get a low score, and an incorrect one with similar notation a high score. The agent optimizes BLEU and is not required to increase correctness; our correctness check is not part of the evaluation scripts, and its results are not reported to the agent. The agent does, however, see the original model answers on dev and may check them itself (Section 7.1). A solution is considered correct if the answer $y(x)$ is parsed by sympy, the number of free constants equals the order of the equation, and the relative residual at five points $x_j$ for random values of the constants $C$ does not exceed $10^{-6}$:

$$\frac{|\,a\,y'' + b\,y' + c\,y - f(x)\,|_{x_j,C}}{|y(x_j;C)| + 1} \le 10^{-6}, \qquad j = 1, \ldots, 5$$

(for first-order equations the residual is $|y' - F(x)|$). The answer is not compared with the reference, so a correct solution in different notation is accepted. For separable, correctness is poorly measured: the model often leaves an integral in the answer, and sympy parses only 18–39% of its answers depending on the split.

**Emotion classification.** The prompt is fixed: the system message “You are a helpful assistant.”, an instruction to choose one of seven emotions and answer with one word, and the comment text; the assistant answer starts with the prefix `The emotion expressed in this comment is "`, which is the form in which the model answers by itself. The answer is a greedy generation of at most 5 tokens; the predicted class is the first word of the answer if it is one of the seven labels (case-insensitive), otherwise the answer belongs to no class. The target metric is macro-F1:

$$F1 = \frac{1}{7}\sum_{c} \frac{2P_cR_c}{P_c + R_c}, \qquad P_c = \frac{TP_c}{\#\{\hat{y} = c\}}, \quad R_c = \frac{TP_c}{\#\{y = c\}},$$

where $TP_c$ is the number of correctly recognized texts of class $c$. An answer outside the label list reduces the recall $R_c$ of the true class. Macro-F1 directly measures classification quality with respect to the reference labels but differs from accuracy (the share of correct answers): averaging over classes accounts for the precision and recall of each class separately. Both quantities are therefore reported. An auxiliary characteristic is the F1 of probability-based label choice: the model is credited with the label among the seven whose first token has the highest probability right after the prefix (summed over lowercase and capitalized spellings, with and without a leading space).

### 5.3. Three tasks

| Task | Objective | Constraints | Form of Δ | Evaluation script |
|---|---|---|---|---|
| Brush (repair) | max $\Delta\mathrm{BLEU}_{\mathrm{ode2}}$ | $\Delta\mathrm{BLEU}_{\mathrm{poly}}, \Delta\mathrm{BLEU}_{\mathrm{sep}} \ge 0$; $\lVert \Delta \rVert_0 \le 131\,072$ | any | `eval_edit.py` |
| Scalpel (pruning) | max $\rho(m)$ | $\Delta\mathrm{BLEU}_t \ge -0.01$ for each type | neuron mask | `eval_prune.py` |
| Layer removal with repair | max number of removed layers | $\Delta\mathrm{BLEU}_t \ge -0.01$; repair $\lVert \Delta \rVert_0 \le 131\,072$ | block removal + any | `eval_layers.py` |

Formally, with $\varepsilon = 0.01$ and $\mathcal{T} = \{\text{ode2}, \text{poly}, \text{sep}\}$:

$$\text{brush: } \max_{\Delta} \Delta\text{BLEU}_{\text{ode2}} \text{ s.t. } \Delta\text{BLEU}_{\text{poly}} \geq 0,\ \Delta\text{BLEU}_{\text{sep}} \geq 0,\ \| \Delta \|_0 \leq 131\,072;$$

$$\text{scalpel: } \max_{m} \rho(m) \text{ s.t. } \Delta\text{BLEU}_t \geq -\varepsilon\ \ \forall t \in \mathcal{T};$$

$$\text{layer removal: } \max_{L,\Delta} |L| \text{ s.t. } \Delta\text{BLEU}_t \geq -\varepsilon\ \ \forall t \in \mathcal{T},\ \| \Delta \|_0 \leq 131\,072,$$

where $L$ is the set of removed layers and $\Delta$ touches only the remaining ones. The limit of 131,072 weights is taken from [1] (64 columns of dimension 1024 in each of two layers). The tolerance $\varepsilon = 0.01$ corresponds to about half a problem out of 50.

These constraints apply during the search on dev. On the held-out splits, a single criterion is used for all three tasks: mean $\Delta\text{BLEU}_t \geq -0.01$ for each type; for the brush it is milder than the search constraint. Borderline cases are decided on unrounded values.

On emotion classification, two tasks of the same kind are posed: the **brush** — maximal $\Delta F1$ of the model's own answer with $\| \Delta \|_0 \leq 131\,072$ (script *eval_cls_edit.py*), and the **scalpel** — the maximal share of pruned MLP neurons with $\Delta F1 \geq -0.01$ for probability-based label choice (script *eval_cls_prune.py*). Probability-based choice is used for the scalpel because the original model's own answer is almost at chance level (Section 7.7), so there is nothing to preserve in it.

### 5.4. Agent and protocol

The **agent** "Sergei" is a digital twin of the researcher who authored the paper. It works in the DeepSeek Harness agent environment (version 0.2.0-rc.2); its engine, i.e. the language model that plans the work, writes code and makes decisions, is **deepseek-v4.1-flash** (the cloud version, called through Ollama). The environment provides tools for working with files and running code and, since 7 November 2026, skills for working with Office documents; the twin's persona and working rules are defined by an instruction file. The agent receives the task description, full access to the project directory (weights, evaluation code, pool, the original model answers on dev), Python with PyTorch and a GPU.

**Rules.**

1. Evaluation on dev only through the script of its task; every call is logged, the limit is 50 calls per run. Generating answers on dev with its own code is forbidden.
2. Anything is allowed on the pool, including fitting edits by gradient optimization.
3. Full fine-tuning is forbidden; in the scalpel task changing the remaining weights is forbidden.
4. Reading the archive of previous runs, our materials (*ours/*, *article/*, controls, our evaluations) and the held-out data is forbidden. Compliance is verified by an audit of the agent's logs.
5. The output is a report *FINAL.md* (result, edit mechanism, tested and rejected hypotheses) and a work journal.

An equal limit means an equal upper bound on the number of dev evaluations, not equal actual consumption: GPU time, computations on the pool, the number of LLM calls and their cost were not equalized.

The **work cycle** is shown in Fig. 1. A run consists of rounds: each round is one DeepSeek Harness session that a control script continues with the message "continue working", up to 30 rounds, until the agent writes *FINAL.md* or exhausts the limit. Within a round the agent plans its work itself. It can offload heavy computations on the pool to a second computer (RTX 3060, 12 GB) through the tool *`tools/home.ps1`*; dev evaluations for ODEs and for the emotion brush are run on a laptop (RTX 3060 Laptop, 6 GB), because greedy generation on another machine gives slightly different answers (Section 9). Each task is run three times; after a run everything the agent left is moved to an archive, and the next run starts from a clean directory. The brush and the scalpel on emotions ran in parallel: the scalpel in a separate working folder, with dev evaluations on the second computer, where dev is kept in a folder closed to the agent; scalpel masks are compared with the original model on that machine.

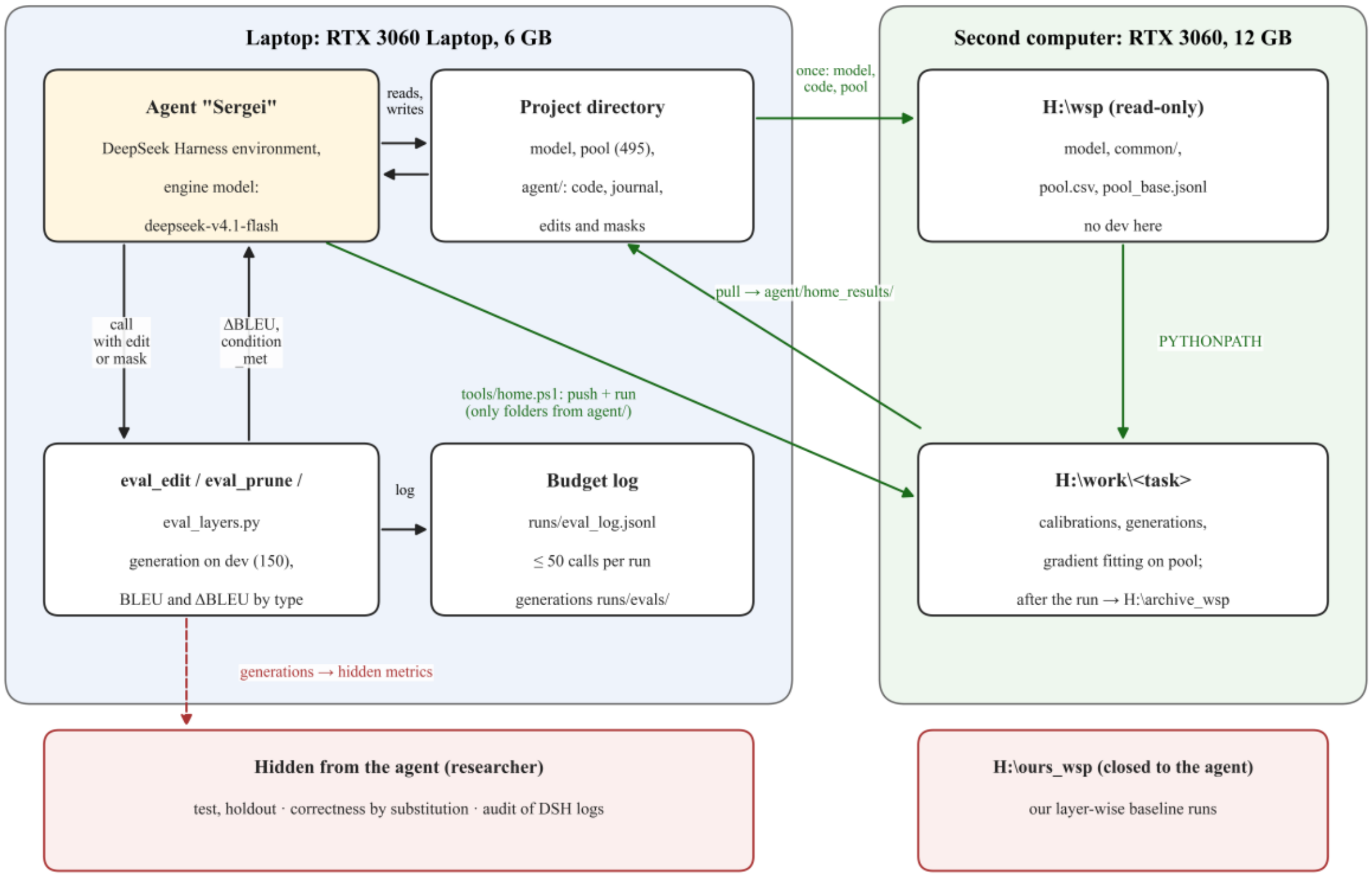

*Fig. 1. The agent's workflow. The agent reads and writes the project directory on the laptop. It evaluates edits and masks on dev only through the* `eval_*.py` *scripts, each call of which is recorded in the budget log. It can offload heavy computations on the pool to the second computer through* `tools/home.ps1`*: push and run send a folder from* `agent/` *to* `H:\work\<task>` *and start it in the background, pull returns the results to* `agent/home_results/`*. The second computer holds the model, code and pool but not dev. The results of our correctness check and the evaluations on test and holdout are not available to the agent. Our baseline measurements on the second computer are kept in a folder closed to the agent.*

## 6. Baselines and Controls

All baselines and controls were computed by us, not by the agent; their purpose is summarized in Table 1.

**Table 1.** Baselines and controls.

| Method or control | Task | What it tests | Results |
|---|---|---|---|
| Random edits of the same size: random columns of the same layers (3 seeds) and the agent's columns with a random direction | brush, ODEs | whether the effect depends on the found direction | Table 2 |

| Method or control | Task | What it tests | Results |
|---|---|---|---|
| Prompt instruction without weight editing: reference style, with examples; on emotions — class definitions, with examples | brush, both domains | whether the effect is achievable without weight editing | Tables 2, 10 |
| α-scan: $W + \alpha\Delta$, $\alpha$ from 0.25 to 1.5 | brush, ODEs | margin of the edit to the threshold | Table 7 |
| Wanda, Taylor and random selection with an equal share in each layer (5, 10, 20%) | scalpel, ODEs | what the agent adds compared with standard pruning | Table 3 |
| ShortGPT and NLL increase on layer removal; 1–4 layers without repair | layer removal | whether repair is needed | Table 5 |
| Random masks of the same share and shape (3 seeds) | scalpel, emotions | whether the choice of neurons matters | Table 11 |
| Ridge and logistic regression written into the weight rows chosen by the agent | brush, emotions | whether a procedure without an LLM reproduces the quality | Table 10 |
| Pruning settings search: three layer-allocation schemes, same evaluation limit | scalpel, emotions | the same for pruning | Table 11 |
| Neuron importance scale (Wanda, Taylor) and Jaccard index of masks | both tasks | where the edited and pruned neurons lie; similarity of runs | Sections 7.3, 8.2 |

The controls answer two different questions. Random edits and masks of the same shape, the α-scan and the prompt instruction show where the effect of the edit found by the agent comes from. Baseline methods of pruning and layer removal, and on emotions the procedures without an LLM, show what the agent itself contributes compared with a standard procedure. Exact definitions, the formulas of the importance scores and the rule for writing a classifier into the weights are given in the Appendix (A.2).

The comparisons are not equal in search freedom. On ODEs, the baseline pruning methods were not given a search over calibration and layer-wise allocation, so the comparison with them does not isolate the contribution of agentic planning. On emotions such a search was carried out: the pruning settings search works with the same constraint and evaluation limit as the agent. The fixed brush procedure, by contrast, uses the weight rows already chosen by the agent and reproduces its solution rather than searching for it from scratch.

## 7. Results

Sections 7.1–7.4 describe what the agent did and what was obtained on dev — the split on which it selected its final variant; these numbers are labeled [dev] and are inflated by the selection itself. In Section 7.5 the final variants are checked on the held-out splits, and Section 7.6 estimates the margin to the threshold — in edit strength and in tolerance; Section 7.7 is devoted to the second domain. The summary table gives the results confirmed on the held-out splits upfront; its numbers are not repeated in the text.

**Summary table.** ΔBLEU is the difference from the original model; the constraint is $\Delta\text{BLEU} \geq -0.01$ for each equation type.

| Task | dev | test (248 problems) | holdout (450 problems) | Correct ode2, test / holdout |
|---|---|---|---|---|
| Brush, 3 runs | ode2 +0.09…+0.13, poly +0.23…+0.52 | ode2 +0.11…+0.15, poly +0.28…+0.53 | ode2 +0.12…+0.17, poly +0.24…+0.54 | 76–83% / 80–85% (original 81% / 83–84%) |
| Scalpel, 12.5 and 15% masks (model smaller by 9.4–11.2%) | constraint ✓ | ✓ | ✓ | 76–79% / 79–90% (81 / 84%) |
| Scalpel, 22% mask (model smaller by 16.5%) | ✓ | ✗ (−0.013), ✓ at tolerance 0.02 | ✓ | 68% / 81% |
| Removal of layer 5 with repair (model smaller by 3.0%) | ✓ in 2 of 3 runs | ✗ (−0.014…−0.015), ✓ at 0.02 | ✗ (−0.012…−0.013), ✓ at 0.02 | 71–72% / 79% |

### 7.1. Brush: repair by weight editing

Three independent runs of the agent produced three different edits (Table 2); each needed 2–3 dev calls out of 50. The BLEU gain relative to the original model is given in parentheses.

**Table 2.** Final brush edits and controls [dev].

| Edit | Where | Weights changed | BLEU ode2 | BLEU poly | BLEU sep | Correct ode2 | Correct poly |
|---|---|---|---|---|---|---|---|
| original model | — | — | 0.461 | 0.363 | 0.116 | 88% | 92% |
| run 1 | 64 `down_proj` columns, layers 15, 19 | 98,304 | **0.590** (+28%) | 0.589 (+62%) | **0.134** | 80% | 92% |
| run 2 | 2 embedding rows | **3,072** | 0.571 (+24%) | **0.886** (+144%) | 0.125 | 84% | **100%** |
| run 3 | 85 `down_proj` columns, layer 27 | 130,335 | 0.551 (+20%) | 0.884 (+143%) | 0.128 | 84% | 96% |
| random columns (3 seeds) | layers 15, 19 | 98,304 | 0.463–0.468 | 0.361–0.364 | 0.108–0.120 | 82–86% | 92% |
| run 1 columns, random direction | layers 15, 19 | 98,304 | 0.457 | 0.359 | 0.112 | 90% | 88% |
| run 3 control (made by the agent) | layer 27 | 130,560 | 0.455 | 0.363 | 0.114 | 84% | 92% |
| original model + notation instruction | prompt | 0 | 0.450 | 0.329 | 0.099 | 82% | 88% |
| + instruction and two | prompt | 0 | 0.462 | 0.327 | 0.091 | 78% | 86% |

| Edit | Where | Weights changed | BLEU ode2 | BLEU poly | BLEU sep | Correct ode2 | Correct poly |
|---|---|---|---|---|---|---|---|
| pool references | | | | | | | |

The edits change 0.0002 to 0.0084% of the model's weights. Random edits of the same size at the same locations change ode2 BLEU by at most about 0.007, and the BLEU of any type by less than 0.009. Hence the effect is determined by the direction found by the agent rather than by the choice of columns.

A prompt instruction does not reproduce the effect of the edit. Asking the model to write the answer in the reference style, alone or with two reference answers, does not raise BLEU either on dev (Table 2) or on test: ode2 BLEU 0.448 and 0.456 versus 0.456 for the original prompt, poly BLEU 0.333 and 0.314 versus 0.356, while the share of correct ode2 solutions drops from 80 to 72–74%. The model does not follow the instruction: none of the 119 ode2 answers on test contains `C_{1}`. The brush on the same test problems raises BLEU of both types (Section 7.5).

**What the agent found.** All three runs started from the same observation: the model writes formulas in a "minimal" style (`C_1`, `x^2`, `\cos(x)`), while the references are in sympy style (`C_{1}`, `x^{2}`, `\cos{\left(x \right)}`); the 150 model answers on the pool contain not a single `^{` or `_{`. The second run checked the answers by substitution into the equation and concluded that the entire loss is due to the notation of the answer rather than the mathematics: the model solves ode2 correctly in 44 cases out of 50. The third run estimated the ceiling: if the model's own answers are rewritten in the reference style, ode2 BLEU rises from 0.461 to 0.834 without any change in the mathematics. The choice "with or without a brace" is the choice of a single next token (`^{` versus `^`, `_{` versus `_`), and the model makes it with a margin of about 20 nats: the braced variant is $e^{20} \approx 5 \cdot 10^8$ times less likely.

**Three implementations of one technique.** Run 1 selected 32 *down_proj* columns in each of layers 15 and 19 with the largest gradient norm and fitted additions to them by gradient descent on the pool; the probability of `_{` after $C$ rose from 0.49 to 0.9997. Run 2 changed two rows of the embedding matrix — those of the tokens `^{` and `_{`; since this matrix also serves as the output layer, editing row $e_t$ shifts the logit of exactly this token. Run 3 built the edit without gradient descent over the model weights: on pool activations it trained a sparse linear detector $v$ on 85 neurons of layer 27 that separates the states "a brace choice comes next" from the rest (AUC 1.000, false positives 0.3%), and wrote a single direction into their *down_proj* columns

$$\Delta W_{\text{down}}[:,j] = c_j\, u, \qquad u = \left(e_{t_1} - e_{t_2}\right) + 0.6\left(e_{t_3} - e_{t_4}\right), \qquad c_j \propto v_j,$$

where $t_1$ = `^{`, $t_2$ = `^`, $t_3$ = `_{`, $t_4$ = `_`. The MLP output receives the addition $u\,(a \cdot v)$ — a steering vector [5–7] written into the weights and switched on only where the detector fires. The same idea underlies SteerEdit [34], but there the activation condition is set by projecting the edit onto a subspace that leaves activations on ordinary inputs untouched, whereas here it is a linear detector trained by the agent itself.

**Correctness control.** On dev no edit increased the share of correct ode2 solutions; it even decreased slightly (80–84% versus 88%; 82–90% for random edits), and for poly it changed by 0–4 problems out of 50. The growth of poly BLEU is explained by rewriting already correct answers into the reference form: for run 2, after the edit 28 of 50 poly answers coincide with the reference verbatim. The picture on the held-out splits is the same (Section 7.5).

### 7.2. Baseline pruning

**Table 3.** Equal share of neurons in each layer [dev].

| Method, share | BLEU ode2 | BLEU poly | BLEU sep | Correct ode2 | Correct poly |
|---|---|---|---|---|---|
| original model | 0.461 | 0.363 | 0.116 | 88% | 92% |
| Wanda 5% | 0.422 | 0.375 | 0.107 | 58% | 94% |
| Taylor 5% | 0.441 | 0.405 | 0.099 | 82% | 98% |
| random 5% | 0.436 | 0.385 | 0.097 | 48% | 90% |
| Wanda 10% | 0.395 | 0.407 | 0.073 | 58% | 94% |
| Taylor 10% | 0.434 | 0.381 | 0.101 | 68% | 96% |
| random 10% | 0.314 | 0.367 | 0.102 | 22% | 64% |
| Wanda 20% | 0.391 | 0.412 | 0.098 | 44% | 90% |
| Taylor 20% | 0.399 | 0.386 | 0.085 | 48% | 94% |
| random 20% | 0.016 | 0.226 | 0.090 | 0% | 18% |

No baseline method satisfied the scalpel constraint at any share: on ode2 and separable BLEU drops by more than 0.01. This result refers to the tested shares and the uniform layer-wise allocation.

### 7.3. Scalpel: pruning by the agent

The result of each run is a scalpel mask (Section 5.1). A pruned neuron removes $3 \cdot 1536 = 4608$ parameters; the original model contains 1,543,714,304 parameters (2,944 MiB in bf16).

**Table 4.** Final masks of the three runs [dev]. Layer groups: 0–6 / 7–13 / 14–20 / 21–27. The course of each run call by call is given in the Appendix (A.3).

| Run | Neurons pruned | Model smaller by | Ranking and layer-wise allocation | BLEU ode2 / poly / sep | Correct ode2 / poly | Dev calls |
|---|---|---|---|---|---|---|
| original | 0 | — | — | 0.461 / 0.363 / 0.116 | 88% / 92% | — |
| r1 | **22%** (55,195) | **16.5%** (485 MiB) | Wanda within layer; 26 / 22 / 14 / 26% | 0.457 / 0.400 / 0.111 | 72% / 100% | 8 |
| r2 | 15% (37,632) | 11.2% (331 MiB) | energy $\overline{a_i^2}\ \|d_i\|^2$ within layer; 0 / 5 / 14 / 41% | 0.452 / 0.379 / 0.131 | 84% / 96% | 6 |
| r3 | 12.5% (31,360) | 9.4% (276 MiB) | Wanda global; 58% of neurons of layers 0–5 removed | 0.463 / 0.365 / 0.117 | 86% / 94% | 6 |

All three runs satisfied the constraint while pruning 12.5–22% of MLP neurons, whereas the baselines failed it even at 5%. According to the logs, the difference is due to three decisions of the agent. *Calibration on the whole answer:* the agent averaged activations over the entire text of the model's own answer; r1's uniform 20% mask overlaps with our Wanda 20% by 80% (Jaccard index) but gives 70% correct ode2 solutions versus 44%. *Within-layer ranking:* activations grow with depth by a factor of about 10, so a global ranking prunes mostly early layers (at 30% — 98% of the neurons of layers 0–3); r1 and r2 switched to within-layer ranking and chose the layer-wise allocation by measuring on the pool the cost of pruning each group of layers, while r3 kept the global ranking but limited the share. *Search for the share boundary:* r1 and r2 searched for it by bisection on dev, r3 on 300 pool problems.

The runs diverge strongly. Their conclusions about “expensive” layers contradict each other, and the r2 and r3 masks hardly overlap: 86 common neurons, Jaccard index 0.00125. One mask prunes neurons mainly in the last layers, the other in the first, while ode2 correctness on dev is 84–86% for both versus 88% for the original model. Hence 12.5–15% of MLP neurons can be removed in different ways with almost no loss of correctness on dev. The 22% mask of r1 passes the BLEU constraint but loses 16 pp of ode2 correctness. No general critical threshold follows from this: share and neuron composition change simultaneously from mask to mask. The agent itself pointed out the risk of overfitting to dev: r1, with a margin below 0.007, proposed a more robust 19% mask, and r2 found that a margin of 0.001 on dev does not hold on the pool. The log audit revealed no access to forbidden materials. The evaluation of the masks on the held-out splits is given in Section 7.5; the position of the pruned neurons on the common importance scale and their relation to the brush neurons are discussed in Section 8.2 (Fig. 2).

### 7.4. Layer removal with repair

The task is to remove as many whole blocks as possible so that BLEU of each type decreases by at most 0.01; after removal, repair of up to 131,072 weights is allowed — the same tool as the brush, instead of the LoRA fine-tuning usually applied after layer removal [20]. Removing one block shrinks the model by 46.8 M parameters, i.e. by 3.0%.

**Table 5.** Layer removal: baselines without repair and final agent variants [dev]. The baselines were computed on the second computer and the agent runs on the laptop; ΔBLEU is measured from the original model on the same machine (Section 9), whose share of correct ode2 solutions is 80 and 88%, respectively.

| Removed layers | Method | Repair | ΔBLEU ode2 / poly / sep | Constraint | Correct ode2 / poly |
|---|---|---|---|---|---|
| 5 | NLL increase | none | +0.001 / +0.010 / +0.005 | ✓ | 74% / 94% |
| 5, 17 | NLL increase | none | −0.077 / +0.056 / −0.025 | ✗ | 26% / 74% |
| 15, 17 | ShortGPT | none | −0.154 / +0.028 / −0.032 | ✗ | 6% / 52% |
| 5 | agent r1 | none | −0.0105 / +0.010 / +0.001 | ✗ | 70% / 98% |
| 5 | agent r1 | on reference answers, 88,064 weights | −0.179 / +0.462 / +0.135 | ✗ | 0% / 60% |
| **5** | agent r2 | **RMSNorm gains, 68,697 weights** | −0.005 / +0.004 / −0.005 | ✓ | 70% / 96% |
| **5** | agent r3 | **85 `down_proj` columns of layer 4, 130,560 weights** | +0.004 / +0.012 / −0.008 | ✓ | 76% / 100% |

One layer can be removed, but not two: already a pair of layers lowers ode2 BLEU by 0.08–0.15 under any criterion (ShortGPT is worse than the NLL increase here — it selects the adjacent layers 15–17), and without repair on the laptop even layer 5 fails, missing the threshold by 0.0005. All three agent runs concluded that layer 5 is the cheapest to remove and that layers 0 and 1 must not be removed, and they differed in the repair method. Run r1 ended with a negative result: repair fitted on the pool reference answers brings the notation to the reference style and raises poly and separable BLEU but drives ode2 correctness to zero. Run r2 fitted the normalization gains of the remaining blocks by self-distillation — in the agent's explanation, they set how strongly a block writes into the shifted residual stream; r3 placed the repair where the removed block used to write — into `down_proj` columns of layer 4. Both variants pass the constraint on dev, but their share of correct ode2 solutions is 70–76% versus 88%: the repair restores the notation of the answer but hardly restores the solution. On the held-out splits this result was not confirmed at a tolerance of 0.01 (Sections 7.5, 7.6). In our search, fewer layers could be removed than from the larger models in [19, 20], although the tasks and quality criteria there are different.

### 7.5. Evaluation on held-out splits

The numbers of Sections 7.1–7.4 were obtained on dev — the same split on which the agent selected its final variant — and are therefore inflated: among several variants that satisfied the constraint, the agent kept the best one, and with it the noise favorable for dev. To estimate the result without this selection, we ran all final variants once on two splits the agent had not seen (Section 3): test (119 ode2, 60 poly, 69 separable problems) and then holdout (150 problems of each type). One ode2 problem amounts to 0.8 pp on test and 0.7 pp on holdout.

The BLEU difference from the original model is computed over pairs "same example — two models". A 95% bootstrap interval is built for it: 5000 times, the same number of problems is resampled with replacement, the difference is recomputed, and the 2.5th and 97.5th percentiles serve as bounds. If the interval contains zero, the difference from the original model is not statistically significant, which does not yet prove that the models are equivalent. Intervals are reported where a conclusion depends on them. To speed things up, holdout was computed on two computers (laptop — original model and brush; second computer — original model, scalpel and layer removal); since greedy generation differs slightly between the two machines (Section 9), each variant is compared with the original model on its own machine. The original model behaves almost identically on all splits: ode2 BLEU 0.454–0.461, poly 0.355–0.363, correct ode2 solutions 81–88%.

The separable references in the original holdout file were printed with sympy (`\frac{\operatorname{atan}{\left(2 x \right)}}{\operatorname{acos}{\left(x \right)}}`), whereas on dev and test they are in the data generator's style (`\arctg(2 \cdot x) \cdot \frac{1}{\arccos(x)}`). Since BLEU compares notation, we rewrote the separable holdout references in the dev and test style: each was parsed with sympy and printed again in the form "$A \cdot B$" or "$A \cdot \frac{1}{B}$". All 150 new references are mathematically identical to the old ones; the same transformation applied to the dev and pool references reproduces their text with BLEU 0.89 (only the order of factors differs). The model's answers did not change — BLEU was recomputed on the stored generations, and the previous values are given below, in item 5. This recomputation shows the sensitivity of the result to the notation of the reference and is not a new independent evaluation of the model. Correctness does not depend on the notation of the reference. For separable it is low on all splits (0–6% for the original model): the model writes only 18–39% of its answers in a closed form that sympy can parse and leaves the rest with an integral or truncated.

**Table 6.** Final variants on the held-out splits: ΔBLEU ode2 / poly / sep and satisfaction of the constraint ΔBLEU ≥ −0.01 for each type [test, holdout].

| Variant | Model smaller by | test: ΔBLEU | test | holdout: ΔBLEU | holdout | Correct ode2, test / holdout |
|---|---|---|---|---|---|---|
| original model | — | 0 / 0 / 0 | — | 0 / 0 / 0 | — | 81% / 83–84% |
| brush r1 | — | **+0.145** / +0.276 / +0.006 | ✓ | **+0.173** / +0.241 / −0.010 | ✓ | 76% / 80% |
| brush r2 | — | +0.133 / **+0.531** / +0.025 | ✓ | +0.139 / **+0.537** / +0.034 | ✓ | 82% / 84% |
| brush r3 | — | +0.110 / +0.534 / +0.023 | ✓ | +0.116 / +0.536 / +0.023 | ✓ | 83% / 85% |
| scalpel r1, 22% | 16.5% | −0.013 / +0.028 / −0.013 | ✗ | +0.004 / +0.017 / −0.003 | ✓ | 68% / 81% |
| scalpel r2, 15% | 11.2% | −0.003 / +0.028 / +0.016 | ✓ | −0.000 / +0.024 / −0.003 | ✓ | 76% / 79% |
| scalpel r3, 12.5% | 9.4% | +0.000 / −0.003 / +0.004 | ✓ | +0.011 / −0.010 / −0.003 | ✓ | 79% / 90% |
| layer 5 without repair | 3.0% | −0.013 / +0.021 / −0.020 | ✗ | −0.005 / +0.011 / −0.010 | ✗ | 64% / 72% |
| layer 5 + repair r2 (normalizations) | 3.0% | −0.015 / +0.007 / −0.008 | ✗ | +0.010 / +0.000 / −0.013 | ✗ | 72% / 79% |
| layer 5 + repair r3 (layer 4 columns) | 3.0% | −0.014 / +0.017 / +0.003 | ✗ | +0.007 / +0.015 / −0.012 | ✗ | 71% / 79% |

Five conclusions follow from Table 6.

1. **The BLEU gain after the brush is confirmed on new problems.** On both splits the gains in ode2 and poly BLEU are of the same order as on dev, the 95% paired bootstrap intervals of the ode2 and poly gains lie above zero, and the share of correct ode2 solutions did not increase: 76–83% on test against 81% for the original model, so a decrease of up to about 5 pp is not excluded (summary table). The edits are therefore not fitted to dev: they change the notation of the answer on new problems as well, while the model did not get better at solving equations.
2. **The 12.5 and 15% scalpel masks pass the criterion** on both splits; the share of correct ode2 solutions changes within −5…+6 pp. The r1 mask (22%) satisfies the constraint on holdout but loses 0.013 on ode2 and separable on test. The intervals of these losses ([−0.030; +0.004] and [−0.042; +0.016]) include zero, i.e. the BLEU drop is not statistically significant, but it is not excluded either; the share of correct ode2 solutions at 22% falls on test as it does on dev. Satisfying the tolerance refers to mean BLEU values and does not mean statistically proven preservation of correctness. For the 22% mask the lower bound of this interval on test, −0.030, lies below the tolerance, so a loss larger than 0.01 is not excluded there.
3. **Layer removal is not confirmed on either split, although it misses the threshold by thousandths** (Table 6). The repair reduces the worst loss on test, does not change it on holdout, and restores the share of correct ode2 solutions by 7–8 pp. The result “one layer removed” rested on a margin of less than one dev problem, and on new problems this margin was not enough.

4. **Variants that passed dev with a minimal margin are not confirmed.** On test, three of the eight final variants of the agent failed, on holdout two, and all of them had passed dev with a margin of less than one problem. Variants with a larger margin — the brush and the scalpel up to 15% — were confirmed on both splits.
5. **The outcome of the check depends on the notation of the reference.** With the separable holdout references in sympy style, the 22% mask lost 0.059 BLEU on separable, layer removal 0.023–0.038, and the brush gained +0.02…+0.09; after bringing the references to the dev and test style, the figures are −0.003, −0.010…−0.013 and −0.01…+0.03, respectively, for the same model answers. This is direct confirmation that BLEU measures the form of notation.

### 7.6. Margin to the threshold: edit strength and tolerance

Variants that satisfied the constraint with a small margin are confirmed worse on the held-out splits (Section 7.5). The size of this margin is shown by two experiments: varying the strength of the brush edit and recomputing the already obtained results under relaxed tolerances.

**Edit strength.** The brush edit of run 1 was multiplied by a coefficient $\alpha$: $W + \alpha\Delta$.

**Table 7.** α-scan of the brush edit of run 1 [dev].

| α | BLEU ode2 | BLEU poly | BLEU sep | Correct ode2 | Correct poly |
|---|---|---|---|---|---|
| 0 | 0.461 | 0.363 | 0.116 | 88% | 92% |
| 0.25 | 0.463 | 0.371 | 0.117 | 84% | 92% |
| 0.5 | 0.500 | 0.373 | 0.119 | 86% | 92% |
| 0.75 | 0.594 | 0.384 | 0.112 | 78% | 94% |
| 1 | 0.590 | 0.589 | 0.134 | 80% | 92% |
| 1.25 | 0.578 | 0.938 | 0.132 | 64% | 92% |
| 1.5 | 0.542 | 0.942 | 0.135 | 40% | 80% |

The edit acts as a switch: ode2 BLEU barely changes up to $\alpha = 0.5$ and jumps to a plateau at $\alpha = 0.75$, when the competing token is outweighed and the greedy answer changes at once; for poly the threshold is higher. The brush has a large margin — the effect holds from $\alpha = 0.75$ to 1.5 — but as the edit strength grows, the solutions deteriorate: at $\alpha = 1.5$ ode2 BLEU is still above the original, while the number of correct solutions is halved.

**Tolerance.** The tolerance $\varepsilon = 0.01$ was chosen in advance, and the agents searched for the boundary for it. We applied tolerances of 0.02, 0.03 and 0.05 to the already computed results without running anything new. For each tolerance, the variant is selected only on dev among all variants already evaluated by the agents and the baselines; the held-out splits are not used in the selection and remain an independent check.

**Table 8.** The largest variant satisfying $\Delta\text{BLEU} \geq -\varepsilon$ for each equation type [dev]. In parentheses: model size reduction and the share of correct ode2 solutions (88% for the original model).

| Tolerance $\varepsilon$ | MLP neurons: agent | MLP neurons: Wanda / Taylor / random | Whole layers |
|---|---|---|---|
| 0.01 | 22% (−16.5%; 72%) | no share | 1 (−3.0%; 70–76%) |
| 0.02 | 25% (−18.7%; 72%) | no share | 1 |
| 0.03 | 25% (−18.7%; 72%) | Taylor 10% (−7.5%; 68%), random 5% (48%) | 1 |

| Tolerance $\varepsilon$ | MLP neurons: agent | MLP neurons: Wanda / Taylor / random | Whole layers |
|---|---|---|---|
| 0.05 | 25% (−18.7%; 72%) | Taylor 10%, Wanda 5% (58%) | 2: layers 5 and 27 with repair (−6.1%; 68%) |

**Table 9.** The smallest tolerance at which the final variants pass on test and holdout (separable on holdout — with references in the dev and test style). Tolerances above 0.01 were applied to variants that had already been computed (Section 7.6).

| Variant | test: min ΔBLEU | required $\varepsilon$ | holdout: min ΔBLEU | required $\varepsilon$ |
|---|---|---|---|---|
| scalpel r1, 22% (−16.5%) | −0.013 | 0.02 | −0.003 | 0.01 |
| scalpel r2, 15% (−11.2%) | −0.003 | 0.01 | −0.003 | 0.01 |
| scalpel r3, 12.5% (−9.4%) | −0.003 | 0.01 | −0.010 | 0.01 |
| layer 5 without repair (−3.0%) | −0.020 | 0.03 | −0.010 | 0.02 |
| layer 5 + repair r2 | −0.015 | 0.02 | −0.013 | 0.02 |
| layer 5 + repair r3 | −0.014 | 0.02 | −0.012 | 0.02 |

At $\varepsilon = 0.02$, the 22% mask and the removal of layer 5 with repair in both runs are confirmed on both held-out splits (Table 9, summary table); without repair the layer needs a tolerance of 0.03 on test, so here repair is needed both for BLEU and for correctness. On dev, under relaxed tolerances the pruning boundary moves only to 25%, the baselines satisfy the constraint only from $\varepsilon = 0.03$ and at a share 2.5 times smaller, and two layers pass only at $\varepsilon = 0.05$, in a single agent variant that was not checked on the held-out splits. The agents hardly tried variants larger than 25% of neurons or two layers, so for relaxed tolerances Table 8 gives a lower bound; the true boundary at $\varepsilon = 0.02$–$0.05$ requires new runs and a new held-out split.

### 7.7. Second domain: emotion classification

**Original model.** On dev, Qwen2.5-Math-1.5B-Instruct answers with a label from the list in only 39% of cases, most often with the word `sympathetic` (154 of 350 answers); its F1 is 0.133 and its accuracy 11.7% — at chance level (1/7 ≈ 14%). The prompt hardly helps: with class definitions F1 is 0.149, with definitions and examples 0.036 (the model starts answering with phrases). Probability-based label choice gives F1 0.20: the model contains information about the emotion, but it does not reach the answer.

**Brush.** The three independent runs used 3, 3 and 5 dev calls out of 50. Runs r1 and r3 changed seven rows of the embedding matrix — one per single-token spelling of each label, 1536 coordinates each (10,752 in total). Since the input embeddings are tied to the output layer, the row of token $c$ determines its logit $z_c = w_c \cdot h$, where $h$ is the final state of the model at the position before the answer; but changing the row also changes $h$ itself if the token occurs in the prompt.

In run r1 the agent trained ridge regression on $h$ on the pool and, for the five tokens that do not occur in the fixed instruction, wrote the edit

$$\delta_c = \alpha A_c - w_c + b_c \kappa u_{\text{const}} + u, \qquad \alpha = 2,$$

where $A_c$ is a row of the linear classifier, $b_c$ is the class calibration, $\kappa u_{\text{const}} \cdot h \approx 1$, and $u$ is a common offset that lifts the labels above the rest of the vocabulary. The label logit becomes $\alpha A_c \cdot h +$ $b_c \kappa u_{\text{const}} \cdot h + u \cdot h$: the model answers with the emotion recognized by its own internal state. The rows of the two remaining labels — disgust and sadness — occur in the instruction; the agent froze their additions at the values of a preliminary edit under which the features were re-collected, so that fitting the other five rows no longer changed the input state. After rounding to bf16, the r1 edit contains 10,740 nonzero values out of the 10,752 coordinates of the seven rows.

Run r3 arrived at the same read-out scheme with a different classifier — linear discriminant analysis (LDA) with shrinkage 0.5. It centered the coefficients and offsets over the seven classes (without this, the common component lifts all logits so much that class differences are lost in bf16 rounding) and wrote them, together with a common offset, into seven rows. The common idea of r1 and r3 is reading out the emotion feature with a linear classifier through the existing output weights; the training method, edit scale and handling of input embeddings differ between the runs.

The label spellings in the final edits are listed below; $\dot{G}$ in tokenizer notation denotes a leading space.

| Label | r1: token ID and spelling | r3: token ID and spelling |
|---|---|---|
| anger | 4003 `anger` | 19234 `Ġanger` |
| disgust | 67062 `Ġdisgust` | 67062 `Ġdisgust` |
| fear | 90371 `Fear` | 41967 `ĠFear` |
| joy | 4123 `joy` | 27138 `ĠJoy` |
| sadness | 50878 `Ġsadness` | 50878 `Ġsadness` |
| surprise | 84484 `ĠSurprise` | 84484 `ĠSurprise` |
| neutral | 59568 `neutral` | 58694 `ĠNeutral` |

In r1 the fixed instruction contains two changed tokens (disgust and sadness), in r3 three (anger, disgust and sadness), so the hidden state of the original model is not preserved in general, and the change of the instruction acts together with the change of the read-out. Run r2 placed the classifier differently — in a 384 × 341 block of the `down_proj` matrix of the last layer (130,944 weights): the edit reads the emotion from 384 MLP activations and adds to the state a vector that raises the logit of the corresponding label. Runs r1 and r3 themselves set up the control "only answer with a label" (the common offset $u$ without a classifier), and r1 also a control with random directions instead of $A$.

**Table 10.** Brush on emotion classification: macro-F1 of the model's own answer [dev, test, holdout; laptop]. Square brackets: 95% bootstrap interval of the difference from the original model (2000 resamples, stratified by class).

| Variant | Edit coordinates | dev | test | holdout | Accuracy, test / holdout | Answers with a label, test |
|---|---|---|---|---|---|---|
| original model | — | 0.133 | 0.100 | 0.175 | 8.4% / 14.1% | 39% |
| prompt: + class definitions | 0 | 0.149 | 0.135 | — | 12.0% / — | 46% |
| only "answer with a label" (r1 control; r3 on dev 0.185) | 10,752 | 0.121 | 0.101 [−0.03; +0.03] | — | 16.7% / — | 100% |
| random directions (r1 control) | 10,752 | 0.092 | 0.103 [−0.03; +0.04] | — | 16.9% / — | 97% |
| **brush r1** (embedding rows) | 10,752 | **0.428** | **0.415** [+0.27; +0.36] | **0.476** [+0.24; +0.35] | 40.8% / 46.9% | 100% |

| Variant | Edit coordinates | dev | test | holdout | Accuracy, test / holdout | Answers with a label, test |
|---|---|---|---|---|---|---|
| **brush r2** (`down_proj`, layer 27) | 130,944 | **0.424** | **0.396** [+0.25; +0.34] | **0.455** [+0.23; +0.34] | 39.4% / 45.0% | 100% |
| **brush r3** (embedding rows) | 10,752 | **0.441** | **0.451** [+0.30; +0.40] | **0.465** [+0.23; +0.35] | 45.3% / 47.9% | 100% |
| no LLM: ridge regression | 10,752 | 0.448 | 0.424 [+0.27; +0.37] | 0.464 [+0.23; +0.34] | 42.0% / 46.2% | 99% |
| no LLM: logistic regression | 10,752 | 0.382 | 0.363 [+0.21; +0.31] | 0.380 [+0.15; +0.26] | 33.1% / 34.0% | 83% |

In all three runs F1 grew by a factor of 2.6–4.5 and accuracy by a factor of 3–5, and the F1 gain on the held-out splits was +0.28…+0.35: the result transfers to new data and is reproduced from run to run. The r1 control, which only forces the model to answer with a valid label, does not raise F1; random directions of the same size in the same rows give nothing either. The analogous r3 control raises F1 on dev from 0.133 to 0.185, versus 0.441 for the full edit. Hence answering with a label may contribute part of the effect, but the main gain requires a meaningful classifier direction; these controls do not give an exact decomposition of the gain. Per-class F1 on holdout for r1 ranges from 0.27 (neutral) to 0.66 (fear); for the original model F1 for anger and sadness was zero. Besides the change in the instruction, the input representation changes if a changed token occurs in the comment itself; for the five r1 tokens not occurring in the instruction, the log reports two such occurrences in 2100 pool texts.

The fixed procedure without an LLM (Section 6) with ridge regression gives 0.424 on test and 0.464 on holdout — within the range of the three agent runs. The run best on test, r3, gives 0.451, 0.027 more, and 0.465 on holdout. Paired intervals between the methods were not computed, so neither superiority nor equivalence is established. Logistic regression gives 0.363 and 0.380: in the raw coordinates of $h$ the norm of its row edit is 6–12 versus 0.6–0.9 for ridge, and small deviations of $h$ during generation in bf16 noticeably shift the logits — consistent with the sensitivity to the edit norm noted by agent r1. The final quality of the brush is thus largely reproduced by a standard procedure. The agent's contribution in this comparison is the independent choice of the location and design of the intervention: the procedure without an LLM uses the rows chosen by r1 and does not reproduce the search for this solution from scratch.

**Scalpel.** The three runs used 10, 7 and 11 dev calls and arrived at different solutions. Run r1 estimated for each neuron how much it shifts the choice among the seven labels (net of the common shift of all seven logits, which does not affect the choice) and pruned the least influential neurons, but no more than 65% in each layer: without such a cap the late layers were pruned almost entirely and rare classes disappeared. The result of r1 is 55% of MLP neurons (137,984), shrinking the model by 41.2% (1,213 MiB in bf16). Run r2 chose a similar criterion but pruned more in late layers and stopped at 14% (35,000 neurons, model smaller by 10.4%): a 14.2% mask already failed the constraint. Run r3 switched off the MLP entirely in six layers — 9, 12, 15, 16, 18 and 21 (21.4%, model smaller by 16.0%). All three runs themselves pointed to dev noise (standard error of the change about 0.02); r3 found its cause: the original model assigns only 4.4% of probability to the seven labels, the choice among them is nearly balanced, and masks of the same share give +0.001 or −0.041 on dev.

**Table 11.** Scalpel on emotion classification [dev, test, holdout; second computer]. Probability-based F1 is the constraint metric; answer F1 is the model's own answer.

| Variant | dev: probability F1 | test: probability F1 | holdout: probability F1 | Answer F1, test / holdout | Answers with a label, test |
|---|---|---|---|---|---|
| original model | 0.190 | 0.179 | 0.197 | 0.099 / 0.167 | 39% |
| **r1, 55%** | 0.215 (+0.025) ✓ | 0.178 (−0.001 [−0.04; +0.04]) ✓ | 0.150 (−0.047 [−0.08; −0.01]) ✗ | 0.044 / 0.051 | 7% |
| **r2, 14%** | 0.193 (+0.002) ✓ | 0.203 (+0.025 [−0.00; +0.06]) ✓ | 0.211 (+0.015 [−0.02; +0.05]) ✓ | 0.076 / 0.120 | 29% |
| **r3, 21% (6 MLP layers)** | 0.191 (+0.001) ✓ | 0.198 (+0.020 [−0.02; +0.06]) ✓ | 0.189 (−0.008 [−0.05; +0.03]) ✓ | 0.015 / 0.009 | 2% |
| random 55% masks, 3 seeds | — | 0.099–0.108 (−0.071…−0.080) | — | 0.000 / — | 0% |
| no LLM: Wanda, global, 15.9% | 0.191 (+0.001) ✓ | 0.175 (−0.004 [−0.04; +0.03]) ✓ | 0.190 (−0.007 [−0.05; +0.03]) ✓ | 0.072 / 0.088 | 18% |

The agent's choice of neurons is much better than random: random masks with the same number of neurons in each layer as the r1 mask drop F1 to chance level, and the model stops naming emotions. However, the scalpel result is unstable: the shares of the three runs differ fourfold (14, 21 and 55%), and only r2 and r3 satisfied the constraint on holdout; for the r1 mask it is violated significantly. Moreover, the model's own answer suffers under all three masks: least for r2, while after r1 and r3 the model names an emotion from the list in only 2–7% of cases instead of 39%; for the r1 mask, probability-based choice coincides with the original model in only 30% of test texts. A constraint on one metric missed a breakdown in another.

The automatic settings search without an LLM (Section 6) used 33 of 50 dev evaluations and selected global ranking at a share of 15.9%; the constraint is satisfied on test and holdout. In terms of share this lies between the results of r2 (14%) and r3 (21.4%). With an equal share in each layer and with a share growing with depth, only 2.5–4% of neurons satisfy the constraint on dev: what matters is the non-uniform layer-wise allocation, which simple global ranking already provides. The search also runs into dev noise (10% fails the constraint, while 15% passes), and the model's own answer after its mask suffers just as after the agent's masks. As on ODEs, a mask selected on a small noisy dev split does not always transfer, and for a model that can hardly solve the task, the quality-preservation constraint restricts little.

## 8. Discussion

### 8.1. What the agent finds by itself and what procedures without an LLM reproduce

The central question of the work is which part of the result belongs to the agent. The answer has two parts.

Without hints and within the evaluation limit, the agent:

- finds out what the target metric measures: all three brush runs on ODEs independently discovered that BLEU depends on the notation of formulas and separated notation from mathematics themselves;
- finds the location and form of the edit: late-layer `down_proj` columns for formula notation; output-layer rows and single-token label words for emotions; in r1 on emotions, an edit norm of about one at which the learned read-out transfers to bf16 generation;

- finds layer-wise pruning allocations substantially better than uniform; at the same time, the successful masks on ODEs are concentrated sometimes in early and sometimes in late layers, so the results give no general rule about which layers to preserve;
- estimates the dev noise and selects variants on the pool;
- sets up controls of the same shape (random direction, "only answer with a label") that separate the sources of the gain.

Procedures without an LLM, once given the solutions found by the agent, reproduce a substantial part of the final quality. On emotions, ridge regression written into the rows chosen by the agent gives F1 within the range of its three runs. The automatic pruning settings search removes a share of neurons lying between the shares of the two agent runs that passed the hidden check (Table 11). These comparisons show that simple procedures are competitive but establish neither statistical equivalence nor superiority of any method. On ODEs no comparable settings search was carried out for the baselines.

Hence the practical role of the agent: it is useful as a researcher who finds where and how to intervene and checks where the effect comes from; the final procedure can then be written without an LLM. Such reproduction, however, is not the same as an independent search for the solution from scratch. A comparison of search with search — in which a standard procedure also chooses the location of the edit by itself — remains the next step.

### 8.2. Pruning and repair in a single coordinate system

The same agent with one tool, one evaluation limit and one target metric solved both tasks: in repair it raised ode2 BLEU 13–19 times more than the best random edit of the same size (Table 2), and in pruning it satisfied a constraint that standard methods with an equal share in each layer did not satisfy (Tables 3, 4). To compare the two operations, we placed the neurons that the agent edited during repair and pruned during pruning on the common Wanda importance scale (A.2): Table 12, Fig. 2.

**Table 12.** Wanda percentile of the neurons edited by the brush and pruned by the scalpel: within the neuron's layer and among all 250,880 neurons (1 is the most important).

| Neurons | Count | Within layer: median | in bottom 20% | Global: median | in bottom 20% |
|---|---|---|---|---|---|
| brush, run 1 | 64 | 0.98 | 0% | 0.95 | 0% |
| brush, run 3 | 85 | 0.55 | 13% | — | — |
| scalpel r1 | 55,195 | 0.11 | 80% | 0.25 | 34% |
| scalpel r2 | 37,632 | 0.16 | 59% | 0.69 | 8% |
| scalpel r3 | 31,360 | 0.34 | 31% | 0.06 | 100% |

The scalpel prunes neurons from the lower part of the scale of their layer; the runs differ in which layers they cut: r1 almost uniformly, r2 mainly in layers 21–27, r3 in layers 0–5. The brush of run 1 modifies neurons near the top of the scale (median 0.98; by Taylor, 92% of them are in the top 10% of their layer), although it selected them by gradient norm, and no scalpel mask overlaps with them. The neurons of the run 3 brush were chosen differently — as features of a linear detector — and are scattered over the scale (median 0.55); the r2 mask, which removes many neurons of layer 27, hits 31 of these 85 neurons. Thus the statement "the brush takes the top of the scale, the scalpel the bottom" holds for an edit found by gradient but not for an edit built through a detector. The global and within-layer scales diverge strongly because activations grow with depth, and "the bottom of the scale" is not a single set of neurons.

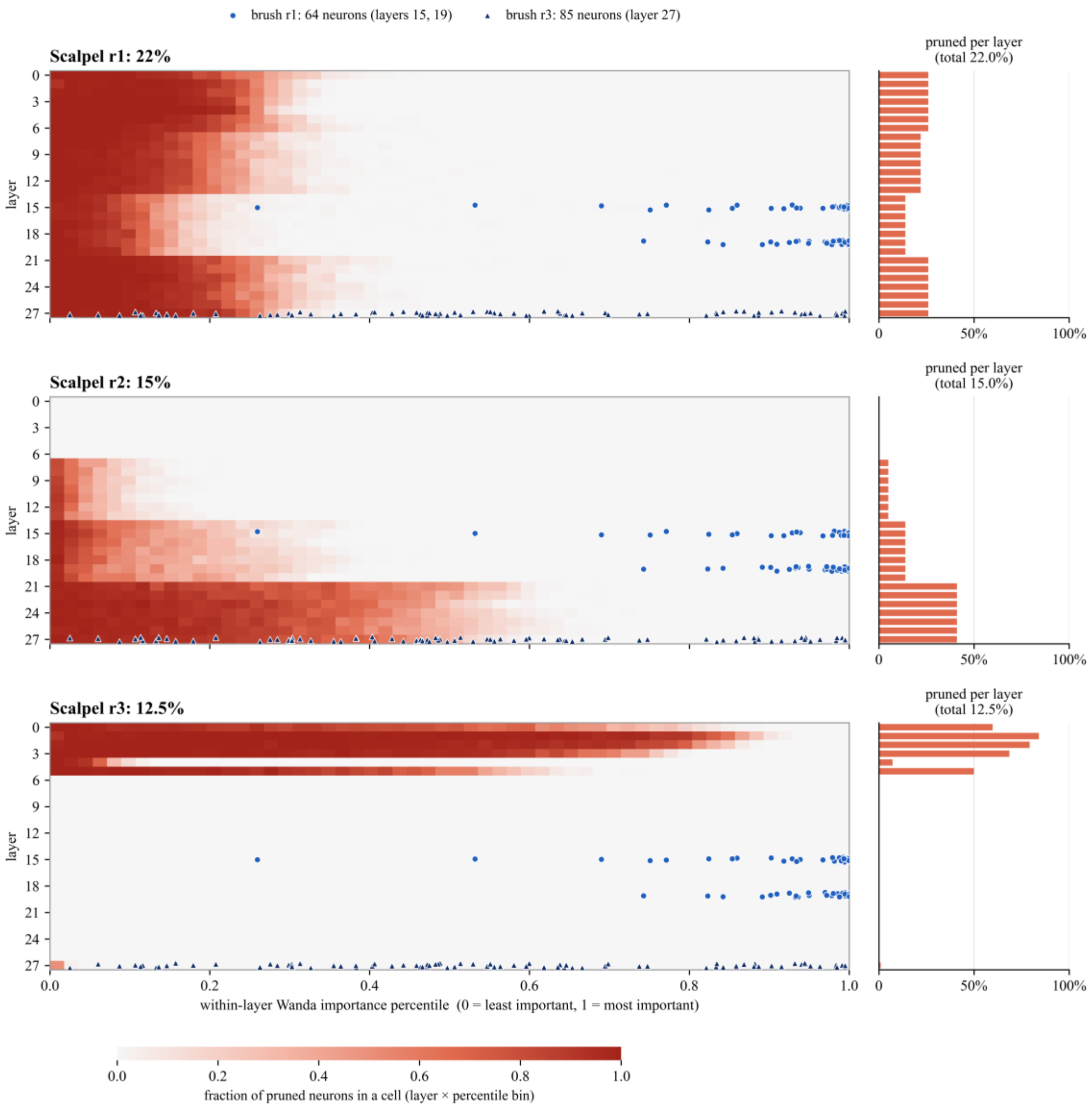

*Fig. 2. Distribution of MLP neurons: what the brush modifies and what the scalpel prunes. Each row is one layer of the model; the neurons of the layer are ordered by Wanda importance from left to right (0 is the least important, 1 the most important). The cell color is the share of pruned neurons in this importance interval (red — pruned, light — kept), one panel per scalpel run. Blue dots are the 64 neurons modified by brush r1 (layers 15 and 19), dark blue triangles are the 85 neurons of brush r3 (layer 27); they are the same on all panels. On the right is the share of pruned neurons per layer. Brush r2 changes embedding rows and is not shown.*

The importance scale thus serves as a common coordinate system in which it is convenient to compare the results of both operations and to check whether they interfere with each other. The common setting should be regarded as an experimental framework for comparing two types of intervention rather than as a single optimal algorithm.

The difference from baseline pruning methods is not due to a new importance score: in the final masks the agent uses a Wanda-like score $\overline{|a_i|}\ \|\ d_i\ \|$. It is due to the decisions around it — calibration on the whole answer, the choice between within-layer and global ranking, and the layer-wise allocation of the share measured on the pool. These techniques can be transferred into an ordinary pruning procedure, but the agent finds them by itself in a few dev evaluations (Table 4). On emotions, where a non-agentic settings search was carried out, it gave a comparable result.

The different shares of the three runs, the contradictory conclusions about "expensive" layers and the almost non-overlapping masks show that the result of an agentic search cannot be reported from a single run: the r2 and r3 masks represent two different ways to compress the MLP. The drop in correctness for the 22% mask on dev and test is consistent with the sharp degradation under pruning described in [25], but testing a phase transition and estimating the critical share requires a series of nested masks built by a single rule.

### 8.3. Target metric and correctness

The agent was given the task of raising BLEU and accomplished it: all three brush runs independently established that BLEU depends on the notation of formulas and raised the metric by changing the notation with a small weight edit (Table 2). The agent was not required to increase mathematical correctness, and the independent check confirmed that it did not change. The sensitivity of BLEU to the notation of the reference is further shown by recomputing holdout with two reference styles for the same model answers (Section 7.5).

BLEU and correctness diverge in both directions. The brush raises BLEU without increasing correctness, and an amplified edit lowers correctness while BLEU stays above the original. The 22% scalpel keeps BLEU but loses a noticeable share of correctness (Table 4); for baseline pruning, already at 5% ode2 BLEU decreases only slightly, while the share of correct solutions for some methods drops almost by half (Table 3). BLEU is suitable when notation must be aligned with a reference, but it guarantees neither the correctness of the solution nor its preservation under compression; if that is what matters, correctness must be included in the objective or the constraints.

On emotions, where macro-F1 directly measures classification quality, the same procedure yielded substantive repair: in all three runs the edit taught the model to name the emotion recognized by its internal state, and F1 grew several-fold on both held-out splits (Table 10). Together with ODEs this shows that the agent raises the metric it is given: in the tasks studied, the growth of BLEU is associated mainly with formula notation, and the growth of macro-F1 with better class choice. A prompt instruction produced no such effect in either domain.

In [1], the edit eliminated looping and was evaluated on hidden tests, including non-mathematical ones. In our experiments on ODEs the mechanism is different — the edit changes the form of notation for the target metric. Random controls and a correctness check independent of the target metric make it possible to distinguish such effects.

### 8.4. Layer removal: pruning and repair together

Layer removal tests both operations at once: first pruning, then repair with the same tool. On this model the combination gave one layer out of 28, and only at a tolerance of 0.02 (Section 7.6); neither the baselines nor the three agent runs found a variant with more layers. This is the limit of the search carried out, not a proof that no better repair exists. Repair on the original model's own answers restores the notation of the answer and only partly its correctness; repair on reference answers raises poly BLEU but drives ode2 correctness to zero (Table 5). The negative result matters for delimiting applicability: at the original tolerance of 0.01, layer removal with repair failed on both held-out splits.

### 8.5. Selection on dev and the role of held-out splits

The results of the brush and of scalpel masks up to 15% transferred to both held-out splits. Variants that satisfied the dev constraint with a margin of less than one problem transferred worse (Section 7.5, Table 6), and on emotions the r1 scalpel mask was not confirmed (Table 11). For such variants, selection on a small dev split introduces a noticeable bias. A held-out split inaccessible to the agent is therefore not an element of reporting but a necessary part of the method.

## 9. Limitations

- **Split size.** The dev split is small: 50 problems per type, one problem is 2 pp; the agent selects its final variant on dev, so the dev estimate is inflated, as confirmed by the held-out evaluation (Section 7.5). The test split is also small (119/60/69 problems), and the ΔBLEU intervals on it have a width of about ±0.017 for ode2. The separable holdout references were brought to the dev and test style only after the first evaluation; BLEU with the original references is given in Section 7.5.
- **Tolerances.** The relaxed tolerances (Section 7.6) were applied to already computed variants; the agents searched for the boundary only for $\varepsilon = 0.01$, so under relaxed tolerances the shares found are lower bounds. Satisfying the tolerance is assessed on means: for some passing variants the confidence interval includes losses larger than 0.01.
- **Model, tasks and agent.** One model and two domains are used; the ode2 and poly references are written in the same style. Correctness for separable is poorly measured: for the original model it is 0–6%, and only 18–39% of its answers are written in a parsable closed form. There is a single agent — DeepSeek Harness with the deepseek-v4.1-flash engine; with other engines the result may differ. The evaluation limit is observed on trust and verified by a log audit.
- **Comparison with baselines.** On ODEs, the baseline pruning methods were tested at 5, 10 and 20% with an equal share per layer, without a search over calibration and allocation, whereas the agent could adapt its strategy on the pool; therefore the difference cannot be attributed unambiguously to the agentic nature of the method. On emotions such a search was carried out (Section 7.7), and no consistent advantage of the agent was found there. The fixed brush procedure uses the weight rows chosen by the agent; whether a standard search would have found them without the agent was not tested. The actual computational costs of the methods were not equalized.
- **Emotions.** Each task was run three times; the brush gave similar results, the scalpel very different ones. The original model recognizes emotions almost at chance level: this leaves a large margin for the brush and makes the quality-preservation constraint weak for the scalpel. The scalpel was checked on probability-based label choice rather than on the model's own answer, and it is the own answer that suffered under all three masks. The held-out splits are taken from an open dataset; the agent was forbidden to download data, and the log audit revealed no violations.
- **Compression.** The reduction in the number of parameters and in weight size is computed from the masks; the speedup of a model with physically removed neurons was not measured.
- **Reproducibility of generation.** Greedy generation is sensitive to the software environment: on the second computer (RTX 3060, a different transformers version), dev answers coincided byte for byte with the original ones in only 70 of 150 problems, with a difference in mean BLEU of at most 0.003. Therefore, within each series of runs the agent's evaluations were performed on a single machine (the emotion scalpel on the second computer), and each variant is compared with the original model on its own machine.

## 10. Conclusion

Repair and pruning of a small language model can be assigned to a single LLM agent with a common representation of interventions $W \to W + \Delta$, where for pruning $\Delta = -W$ on the removed components. This does not require one of the leading language models, as in [1]: a lightweight deepseek-v4.1-flash engine was sufficient for our agent. The main conclusion of the work concerns the agent's contribution. The agent itself finds what the metric measures, where and in what form to edit the weights and how to allocate pruning across layers, and it sets up controls that separate the sources of the effect. In final quality it does not outperform standard procedures without an LLM when they are given the edit location or settings space found by the agent, and borderline variants selected on dev do not always transfer to hidden data.

On ODEs the agent accomplished the assigned task: with a small weight edit it noticeably raised BLEU, bringing the notation of formulas to the reference style, with an unchanged share of correct solutions, and a prompt instruction did not produce the same effect. By pruning, the agent shrank the model while preserving BLEU on both held-out splits, whereas standard methods with an equal share per layer failed the constraint; only one whole layer could be removed with repair, and only under a relaxed tolerance (summary table). On emotion classification, a weight edit raised macro-F1 several-fold on the held-out splits, but procedures without an LLM that were given the solutions found by the agent reached comparable quality (Tables 10, 11).

Practical conclusions follow: the metric given to the agent must match what is actually needed; side effects must be monitored separately, and the result must be checked on hidden data. The next steps are to compare search with search, letting a standard procedure choose the edit location by itself within the same evaluation limit, and to give the agent the task of increasing the correctness of ODE solutions, verified on a new held-out split.

## Appendix. Supplementary materials

### A.1. Related work: details

#### *A.1.1. Localization of behavior and direct editing*

Knowledge-editing work assumes that a specific behavior of a model is localized in a small part of its weights. ROME [2] uses causal tracing (noise is injected into the model and intermediate states are restored one by one to find those on which the answer depends) to show that factual knowledge is stored in the MLPs of middle layers and that a fact can be replaced by a rank-one edit — an addition $uv^\top$ to a single matrix; MEMIT [3] extends the approach to thousands of facts at once. The evaluation methodology of these works is also important for us: an edit is considered successful only if it changes the target behavior and leaves the rest intact, which requires control queries.

Another approach is activation steering. Many behavioral properties (tone, style, refusal tendency) correspond to nearly linear directions in the residual stream: by adding a vector found from the difference of activations on contrastive examples, one can switch a property on and off without changing the weights [5–7]. A steering vector is cheap and reversible, but it acts always rather than only in the relevant context. There is a direct link between the two approaches: column $d_i$ of the MLP output matrix is a fixed direction that the neuron adds to the residual stream in proportion to its activation. Hence editing `down_proj` columns implements a steering vector that is switched on by a condition defined by the neuron's activation. A close construction was proposed by Yin et al. [34]: in SteerEdit a steering vector is written into a low-rank edit of `down_proj` so that it is activated only when a given trigger is present in the input (there this serves as a stealthy backdoor). As shown in Section 7.1, our agent arrives at a similar construction on its own.

The search for an edit can be automated. Malik [4] trains with reinforcement learning a policy that chooses an edit in two modes — multiplying weights by a mask (MaskWorld) and adding a shift (ShiftWorld) — thereby treating mask and shift in a single setting. However, the edit there is found by a specially trained policy, only for classifiers and only for debiasing and unlearning. In [1], weight editing of a language model was performed by another LLM that reads the weights and formulates hypotheses itself. We complement this setting with random edits of the same size and a separate correctness check that distinguishes a change in the notation of the answer from a change in the solution itself.

### *A.1.2. Pruning: importance criteria, calibration and layer-wise allocation*

A pruning method decides which elements of the network to remove: individual weights or whole structures — neurons, channels, attention heads, layers. To this end each element is assigned an **importance score** — a number showing how much the network will degrade without this element; the elements with the lowest score are removed. Optimal Brain Damage [8] estimated the contribution of a weight through the second derivative of the loss; later it turned out that for large networks the weight magnitude followed by fine-tuning also works well [9]. Molchanov et al. [10] proposed a first-order Taylor expansion with respect to activations, which approximates the change in loss from removing a neuron in a single backward pass; SNIP [11] moved the estimate to initialization time. The lottery ticket hypothesis [12] gave pruning a theoretical justification: a trained network contains small subnetworks that preserve quality. Methods with a learned mask (L0 regularization [13], movement pruning [14]) find such subnetworks better but require fine-tuning.

For large language models, fine-tuning after pruning is too expensive, and attention has shifted to methods that need only a small calibration set — a set of examples on which activations are measured. SparseGPT [15] solves a layer-wise reconstruction problem for each layer's output. The authors of Wanda [16] showed that, because of activation outliers in LLMs, weight magnitude poorly reflects a weight's contribution, and proposed the criterion $|W_{ij}| \cdot \| X_j \|$, which is almost as good as SparseGPT. Structured methods remove whole neurons, attention heads and channels and therefore speed up the model without sparse-computation support: LLM-Pruner [17] groups coupled structures and scores them with gradients, SliceGPT [18] rotates the residual-stream basis and deletes rows and columns of matrices. The limiting case is the removal of whole transformer blocks: ShortGPT [19] ranks blocks by how little they change the residual stream, SLEB [21] removes blocks iteratively by their effect on quality, and Gromov et al. [20] showed that a large share of deep layers can be removed with a moderate loss of quality if the model is then restored by QLoRA fine-tuning.

Two conclusions follow from this part of the review. First, the quality of LLM pruning is determined not only by the criterion but also by the calibration data and the layer-wise allocation of the pruning share, and in most works these decisions are made by a human. Second, pruning does not necessarily degrade the model: in SPRINT [22], removing individual attention heads improves reasoning accuracy, including for Qwen2.5-Math-1.5B-Instruct, and Ali et al. [23] found neurons whose removal improves generalization. The boundary between compression and repair is blurred already at the level of individual network components.

### *A.1.3. A physical view of pruning*

The criteria of A.1.2 answer the question of what to remove but not how much can be removed. In our work [24, 25] this question is addressed with statistical physics: the distribution of network activations is treated as an ensemble, and the free energy of activations as a macroscopic quantity reflecting the state of the network. In [24] it is shown that the free energy predicts the accuracy of a network after pruning. In [25] pruning is described as a phase transition: extrema, inflection points and curvature changes of the free-energy profile indicate the critical pruning share beyond which accuracy drops sharply, and the renormalized free energy makes it possible to estimate this threshold from the activations of the original network without repeated runs (up to a 550-fold speedup); the approach was tested on networks ranging from MLP and ResNet to BERT, T5 and GPT-2. This leads to the expectation that agent-driven pruning also has a critical share. The drop in correctness for the agent's largest mask (22%, Sections 7.3 and 7.5) is consistent with this expectation, but it does not prove a general threshold: masks of different shares also differ in neuron composition and layer-wise allocation.

#### *A.1.4. LLM agents as researchers and as a compression tool*

LLM agents are already able to carry out research from hypothesis to report: The AI Scientist [29] automates the cycle up to writing a paper, and MLAgentBench [30] measures how agents cope with machine-learning tasks. In "LLMs can Compress LLMs" [26], an agent assigns layer-wise pruning ratios for Qwen3 using sensitivity profiles, activation and gradient statistics and the history of evaluations. AgenticPruner [27] searches for compression strategies within a given procedure. AutoPruner [28] jointly chooses the layers to remove and the recovery methods for vision-language-action models, relying on functional probes and the history of runs. Agent-driven pruning, including recovery, is thus already being studied. Our agent differs in having open access to weights and activations: it writes its own analysis code and builds sparse edits itself within an evaluation limit. In addition, we test repair, neuron pruning and layer removal on the same model and with the same metric.

### A.2. Baselines and controls: definitions

All baselines and controls were computed by us, not by the agent.

- **Random edits** of the same size as the final brush edit: random columns of the same layers with the same norms and a Gaussian direction (rcol, three seeds) and the agent's columns with a random direction (rdir).

- **α-scan** of the agent's edit: $W + \alpha\Delta$, $\alpha \in \{0.25, 0.5, 0.75, 1, 1.25, 1.5\}$ (Section 7.6).

- **Neuron importance scale.** For each of the 250,880 MLP neurons an importance score is computed — how much the model's answer changes without this neuron; neurons ordered by it form the scale. The scores are computed on 150 pool problems. The model is given the prompt and the last 700 tokens of its own answer (the agent, by contrast, calibrated its scores on the whole answer) in teacher-forcing mode: at each position the next-token distribution given the fixed text is taken. Let $\mathcal{P}$ denote the answer positions and $L$ the mean NLL (negative log-probability of the actual next token):

$$s_i^{\text{act}} = \frac{1}{|\mathcal{P}|}\sum_{t\in\mathcal{P}} |\, a_{t,i}|, \qquad s_i^{\text{Wanda}} = s_i^{\text{act}} \, \| \, d_i \, \|_2,$$

$$s_i^{\text{Taylor}} = \left|\frac{1}{|\mathcal{P}|}\sum_{t\in\mathcal{P}} a_{t,i} \, \frac{\partial L_{\text{own}}}{\partial a_{t,i}}\right|, \qquad s_i^{\text{ref}} = \frac{1}{|\mathcal{P}|}\sum_{t\in\mathcal{P}} a_{t,i} \, \frac{\partial L_{\text{ref}}}{\partial a_{t,i}}.$$

Here $L_{\text{own}}$ is the loss on the model's own answer and $L_{\text{ref}}$ the loss on the reference answer substituted after the model's own reasoning; the first-order Taylor expansion gives $L(a_i = 0) - L \approx -\sum_t a_{t,i} \, \partial L / \, \partial a_{t,i}$. The scales are compared by percentile — the share of neurons with a lower score (0 is the least important, 1 the most important) — within a layer and among all 250,880 neurons.

- **Baseline pruning** on ODEs: Wanda, Taylor and random selection with an equal share of neurons in each layer (5, 10, 20%). Unlike the agent, these variants were not given a search over calibration and layer-wise allocation, so the comparison with them does not isolate the contribution of agentic planning.

- **Baseline layer removal:** block influence by ShortGPT [19], $\mathrm{BI}_\ell = 1 - \frac{1}{|\mathcal{P}|}\sum_t \cos\left(h_{t,\ell}, h_{t,\ell+1}\right)$, and the NLL increase when a single layer is removed, $\Delta L_\ell$; 1–4 layers with the lowest values are removed, without repair.

- **Prompt control** — the original model with an extended prompt and no weight edit. On ODEs — an instruction to write the answer in the reference style (`C_{1}`, `x^{2}`, function arguments in `\left( … \right)`, polynomial in ascending powers, constant at the end) and the same instruction with two reference answers from the pool. On emotions — definitions of the seven classes, and definitions together with one pool example per class.

- **Random masks** for the scalpel on emotions: the same 55% share with the same number of neurons in each layer as the agent's mask, but with randomly chosen neurons (three seeds).

- **Fixed brush procedure without an LLM** (emotions). It uses the same 7 rows of the embedding matrix as run r1 and the final state $h$ of the original model before the answer, collected on the pool; unlike r1, the features are not re-collected after a preliminary edit of the input rows. A standard linear classifier $(A, b)$ is trained on $h$; the regularization parameter is chosen by 5-fold cross-validation on macro-F1 on the pool. The classifier is written into the rows by a predefined rule

$$w_c \to A_c + b_c \frac{m}{m \cdot m} + t \frac{m}{\| m \|},$$

  where $m$ is the mean of $h$ over the pool and $t$ is the smallest common offset at which, on 99% of pool texts, the best of the seven labels exceeds the rest of the vocabulary by at least 1 logit. All 7 rows are changed, including the two whose words appear in the instruction. Two classifier variants were fixed in advance, and both are reported without selection: ridge regression on centered one-hot labels in the raw coordinates of $h$, and multinomial logistic regression on standardized $h$. Neither dev nor the held-out splits are used in the construction.

- **Automatic search over pruning settings** (emotions). Neuron importance is the Wanda score on the pool without labels, $s_i = \mathrm{RMS}_t(a_{t,i}) \ \| d_i \|_2$ over all prompt tokens. Three layer-allocation schemes are searched — an equal share in each layer, global ranking over the whole model, and a share growing linearly with depth — together with the overall share $p$. First, for each scheme the grid $p \in \{0.05, 0.10, 0.15, 0.20, 0.30, 0.40, 0.55\}$ is tested, then the boundary is refined by bisection to an interval below 0.005. The constraint and limit are the same as for the agent: $\Delta F1$ of probability-based choice on dev not below −0.01, at most 50 evaluations. The output is the largest share that satisfies the constraint on dev.

- **Mask similarity** of two runs — the Jaccard index $J(m_1, m_2) = |m_1 \wedge m_2| / |m_1 \vee m_2|$.

### A.3. Course of the scalpel runs

**r1.** Global ranking pruned early layers; the agent switched to within-layer ranking, measured on the pool the cost of pruning each group of layers (the most expensive is 14–20, pruning 21–27 even raises BLEU) and rejected gradient importance, scoring on the end of the answer, calibration on ode2 only, iterative re-scoring and a tilt toward deep layers.

| # | Mask | Share | BLEU ode2 | BLEU poly | BLEU sep | Constraint | Correct ode2 | Correct poly |
|---|---|---|---|---|---|---|---|---|
| 1 | uniform | 0.20 | 0.450 | 0.396 | 0.101 | ✗ | 70% | 98% |
| 2 | uniform | 0.15 | 0.463 | 0.381 | 0.113 | ✓ | 76% | 96% |
| 3 | uniform | 0.18 | 0.467 | 0.388 | 0.135 | ✓ | 74% | 98% |
| 4 | uniform | 0.19 | 0.459 | 0.398 | 0.130 | ✓ | 70% | 100% |
| 5 | **tilt 26/22/14/26% (final)** | **0.22** | **0.457** | **0.400** | **0.111** | ✓ | **72%** | **100%** |
| 6 | tilt | 0.25 | 0.443 | 0.371 | 0.115 | ✗ | 74% | 92% |
| 7 | strong tilt | 0.25 | 0.454 | 0.382 | 0.100 | ✗ | 72% | 98% |
| 8 | medium tilt | 0.24 | 0.443 | 0.406 | 0.104 | ✗ | 78% | 100% |

**r2.** The first call (uniform 24%) collapsed; the agent found on the pool that layers 0–6 are 14 times more expensive per neuron than late ones, built “share → damage” curves per group and optimized the layer-wise allocation.

| # | Mask | Share | BLEU ode2 | BLEU poly | BLEU sep | Constraint | Correct ode2 | Correct poly |
|---|---|---|---|---|---|---|---|---|
| 1 | uniform | 0.24 | 0.265 | 0.355 | 0.031 | ✗ | 8% | 70% |
| 2 | layers 0–6 untouched | 0.22 | 0.431 | 0.377 | 0.103 | ✗ | 60% | 94% |
| 3 | layer-wise allocation | 0.17 | 0.447 | 0.392 | 0.124 | ✗ | 76% | 98% |
| 4 | layer-wise allocation | 0.14 | 0.467 | 0.379 | 0.134 | ✓ | 82% | 96% |
| 5 | layer-wise allocation | 0.16 | 0.445 | 0.379 | 0.130 | ✗ | 70% | 94% |
| 6 | **allocation 0/5/14/41% (final)** | **0.15** | **0.452** | **0.379** | **0.131** | ✓ | **84%** | **96%** |

**r3.** Global Wanda ranking; the boundary was searched on 300 pool problems, dev was used for verification.

| # | Mask | Share | BLEU ode2 | BLEU poly | BLEU sep | Constraint | Correct ode2 | Correct poly |
|---|---|---|---|---|---|---|---|---|
| 1 | global | 0.15 | 0.456 | 0.360 | 0.100 | ✗ | 86% | 94% |
| 2 | global | 0.10 | 0.467 | 0.362 | 0.120 | ✓ | 82% | 94% |
| 3 | **global (final)** | **0.125** | **0.463** | **0.365** | **0.117** | ✓ | **86%** | **94%** |
| 4 | within layer | 0.125 | 0.445 | 0.371 | 0.122 | ✗ | 84% | 94% |
| 5 | global | 0.14 | 0.452 | 0.359 | 0.103 | ✗ | 82% | 90% |

| # | Mask | Share | BLEU ode2 | BLEU poly | BLEU sep | Constraint | Correct ode2 | Correct poly |
|---|---|---|---|---|---|---|---|---|
| 6 | global | 0.11 | 0.467 | 0.366 | 0.107 | ✓ | 82% | 92% |

## References

1. Sparks of Weight-Space Programming. Blog post, 2026. https://rikiyat.github.io/blog/weight-space-programming/
2. Meng K., Bau D., Andonian A., Belinkov Y. Locating and Editing Factual Associations in GPT. arXiv:2202.05262.
3. Meng K., Sen Sharma A., Andonian A. et al. Mass-Editing Memory in a Transformer. arXiv:2210.07229.
4. Malik S. Reinforcement Learning for Neural Model Editing. arXiv:2606.13461.
5. Turner A. M., Thiergart L., Leech G. et al. Steering Language Models With Activation Engineering. arXiv:2308.10248.
6. Zou A., Phan L., Chen S. et al. Representation Engineering: A Top-Down Approach to AI Transparency. arXiv:2310.01405.
7. Panickssery N., Gabrieli N., Schulz J. et al. Steering Llama 2 via Contrastive Activation Addition. arXiv:2312.06681.
8. LeCun Y., Denker J. S., Solla S. A. Optimal Brain Damage. NeurIPS 1989.
9. Han S., Pool J., Tran J., Dally W. Learning both Weights and Connections for Efficient Neural Networks. arXiv:1506.02626.
10. Molchanov P., Tyree S., Karras T. et al. Pruning Convolutional Neural Networks for Resource Efficient Inference. arXiv:1611.06440.
11. Lee N., Ajanthan T., Torr P. H. S. SNIP: Single-shot Network Pruning based on Connection Sensitivity. arXiv:1810.02340.
12. Frankle J., Carbin M. The Lottery Ticket Hypothesis. arXiv:1803.03635.
13. Louizos C., Welling M., Kingma D. P. Learning Sparse Neural Networks through L0 Regularization. arXiv:1712.01312.
14. Sanh V., Wolf T., Rush A. M. Movement Pruning: Adaptive Sparsity by Fine-Tuning. arXiv:2005.07683.
15. Frantar E., Alistarh D. SparseGPT. arXiv:2301.00774.
16. Sun M., Liu Z., Bair A., Kolter J. Z. A Simple and Effective Pruning Approach for Large Language Models. arXiv:2306.11695.
17. Ma X., Fang G., Wang X. LLM-Pruner: On the Structural Pruning of Large Language Models. arXiv:2305.11627.
18. Ashkboos S., Croci M. L., Gennari do Nascimento M. et al. SliceGPT: Compress Large Language Models by Deleting Rows and Columns. arXiv:2401.15024.
19. Men X., Xu M., Zhang Q. et al. ShortGPT: Layers in Large Language Models are More Redundant Than You Expect. arXiv:2403.03853.
20. Gromov A., Tirumala K., Shapourian H. et al. The Unreasonable Ineffectiveness of the Deeper Layers. arXiv:2403.17887.
21. Song J., Oh K., Kim T. et al. SLEB: Streamlining LLMs through Redundancy Verification and Elimination of Transformer Blocks. arXiv:2402.09025.
22. Nguyen H. T., Nguyen B., Nguyen V. A. Structured Pruning for Diverse Best-of-N Reasoning Optimization. arXiv:2506.03978.

23. Ali A., Katz S., Wolf L. Detecting and Pruning Prominent but Detrimental Neurons in Large Language Models. arXiv:2507.09185.
24. Surkov A., Koltcov S., Ignatenko V., Mehmood R., Kupitman K. Free energy of neural network can predict accuracy after pruning // Physica A: Statistical Mechanics and its Applications. 2026. Vol. 681. Art. 131085. DOI: 10.1016/j.physa.2025.131085.
25. Mehmood R., Koltcov S., Surkov A., Ignatenko V. Modeling Pruning as a Phase Transition: A Thermodynamic Analysis of Neural Activations // Computers, Materials & Continua. 2026. Vol. 86, No. 3. Art. 99. DOI: 10.32604/cmc.2025.072735.
26. Kodathala S. V., Vunnam R. LLMs can Compress LLMs: Adaptive Pruning by Agents. arXiv:2601.09694.
27. Esmat S., Banisharif M., Jannesari A. AgenticPruner. arXiv:2601.12272.
28. Lin T. et al. AutoPruner: Agentic Depth Pruning for Vision-Language-Action Models. arXiv:2610.23880.
29. Lu C., Lu C., Lange R. T. et al. The AI Scientist: Towards Fully Automated Open-Ended Scientific Discovery. arXiv:2408.06292.
30. Huang Q., Vora J., Liang P., Leskovec J. MLAgentBench: Evaluating Language Agents on Machine Learning Experimentation. arXiv:2310.03302.
31. Hu E. J. et al. LoRA: Low-Rank Adaptation of Large Language Models. arXiv:2106.09685.
32. Yang A., Zhang B., Hui B. et al. Qwen2.5-Math Technical Report: Toward Mathematical Expert Model via Self-Improvement. arXiv:2409.12122.
33. Papineni K., Roukos S., Ward T., Zhu W.-J. BLEU: a Method for Automatic Evaluation of Machine Translation // Proc. ACL 2002. P. 311–318. DOI: 10.3115/1073083.1073135.
34. Yin R., Han T., Xu N. et al. Compiling Activation Steering into Weights via Null-Space Constraints for Stealthy Backdoors // Proc. ACL 2026 (Main Conference). arXiv:2604.12359.
35. Demszky D., Movshovitz-Attias D., Ko J., Cowen A., Nemade G., Ravi S. GoEmotions: A Dataset of Fine-Grained Emotions // Proc. ACL 2020. P. 4040–4054. DOI: 10.18653/v1/2020.acl-main.372.
36. MathSteering: project repository. https://github.com/hse-scila/MathSteering